\documentclass[sigconf]{acmart}

\AtBeginDocument{%
  }

\copyrightyear{2025}
\acmYear{2025}
\setcopyright{acmlicensed}
\acmConference[MM '25]{Proceedings of the 33rd ACM International Conference on Multimedia}{October 27--31, 2025}{Dublin, Ireland}
\acmBooktitle{Proceedings of the 33rd ACM International Conference on Multimedia (MM '25), October 27--31, 2025, Dublin, Ireland}
\acmDOI{10.1145/3746027.uuuuuuu}
\acmISBN{979-8-4007-uuuu-2/2025/10}

\usepackage{multirow}
\usepackage{multicol}
\usepackage{tabularx}

\usepackage{subcaption}

\usepackage{pifont}

\usepackage[most]{tcolorbox}

\usepackage{amsmath}  
\usepackage{bm}       
\usepackage{enumitem}

\usepackage{graphicx}

\newcommand{\ul}[1]{\underline{#1}}

\begin{document}


\title[DART: Dual Adaptive Refinement Transfer for Open-Vocabulary Multi-Label Recognition]{DART: Dual Adaptive Refinement Transfer for Open-Vocabulary Multi-Label Recognition}

\author{Haijing Liu}
\affiliation{%
  \institution{Sun Yat-sen University}
  \city{Guangzhou}
  \country{China}
  }
\email{liuhj66@mail2.sysu.edu.cn}

\author{Tao Pu}
\affiliation{%
  \institution{Sun Yat-sen University}
  \city{Guangzhou}
  \country{China}
  }
\email{putao3@mail2.sysu.edu.cn}

\author{Hefeng Wu$^{\ast}$}
\affiliation{%
  \institution{Sun Yat-sen University}
  \city{Guangzhou}
  \country{China}
  }
\email{wuhefeng@gmail.com}
\thanks{$^{\ast}$ Corresponding authors}

\author{Keze Wang}
\affiliation{%
  \institution{Sun Yat-sen University}
  \city{Guangzhou}
  \country{China}
  }
\email{kezewang@gmail.com}

\author{Liang Lin$^{\ast}$}
\affiliation{%
  \institution{Sun Yat-sen University}
  \city{Guangzhou}
  \country{China}
  }
\email{linliang@ieee.org}

\renewcommand{\shortauthors}{Haijing Liu, Tao Pu, Hefeng Wu, Keze Wang, Liang Lin}

\begin{abstract}

Open-Vocabulary Multi-Label Recognition (OV-MLR) aims to identify multiple seen and unseen object categories within an image, requiring both precise intra-class localization to pinpoint objects and effective inter-class reasoning to model complex category dependencies. While Vision-Language Pre-training (VLP) models offer a strong open-vocabulary foundation, they often struggle with fine-grained localization under weak supervision and typically fail to explicitly leverage structured relational knowledge beyond basic semantics, limiting performance especially for unseen classes. To overcome these limitations, we propose the Dual Adaptive Refinement Transfer (DART) framework. DART enhances a frozen VLP backbone via two synergistic adaptive modules. For intra-class refinement, an Adaptive Refinement Module (ARM) refines patch features adaptively, coupled with a novel Weakly Supervised Patch Selecting (WPS) loss that enables discriminative localization using only image-level labels. Concurrently, for inter-class transfer, an Adaptive Transfer Module (ATM) leverages a Class Relationship Graph (CRG), constructed using structured knowledge mined from a Large Language Model (LLM), and employs graph attention network to adaptively transfer relational information between class representations. DART is the first framework, to our knowledge, to explicitly integrate external LLM-derived relational knowledge for adaptive inter-class transfer while simultaneously performing adaptive intra-class refinement under weak supervision for OV-MLR. Extensive experiments on challenging benchmarks demonstrate that our DART achieves new state-of-the-art performance, validating its effectiveness.

\end{abstract}



\maketitle




\section{Introduction}

Multi-Label Recognition (MLR) aims to identify the multitude of semantic concepts present in everyday images~\cite{ssgrl, scenegnn, chen2019multi, PuLWCTLL24tmm, wang2017multi, sim-mlr}. However, traditional MLR methods are typically confined to predefined label sets, limiting their applicability when encountering objects from previously unseen categories. Addressing this crucial gap, Open-Vocabulary Multi-Label Recognition (OV-MLR)~\cite{mkt,LiuPWWL24ArXiv} has emerged, focusing on recognizing both seen and unseen classes within an image. This ability to generalize to novel concepts is vital for real-world systems like autonomous driving~\cite{autodrive1, autodrive2}, dynamic scene understanding~\cite{scene1, scene2, PuCWLL24tip}, and large-scale content annotation~\cite{social1, social2}, where new object categories appear frequently.

\begin{figure}[!t]
    \centering
    \includegraphics[width=0.97\linewidth]{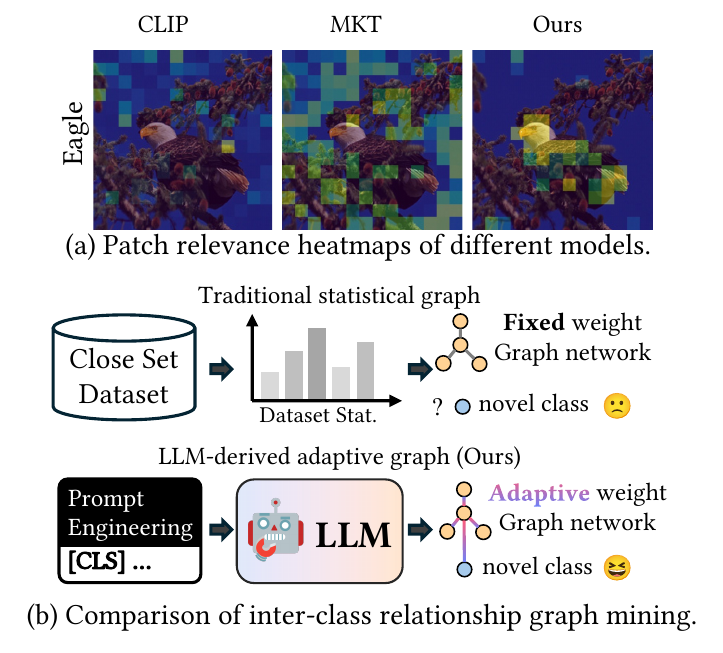}
    \caption{Addressing localization and relational reasoning challenges in OV-MLR.}
    \Description{}
    \label{fig:intro}
\end{figure}

Recent progress in vision-language pre-training (VLP), exemplified by models like CLIP~\cite{clip}, has significantly advanced open-vocabulary tasks by learning aligned embeddings from vast image-text data. Consequently, adapting VLPs has become a dominant approach for OV-MLR~\cite{mkt, lin2024tagclip}. A common strategy involves leveraging patch-level features from the VLP's vision encoder to capture finer-grained visual details~\cite{mkt,lin2024tagclip,WuCLCCL24tcsv}. For instance, methods like MKT~\cite{mkt} attempt to derive class-specific visual representations by computing patch-class similarity scores and selecting relevant patch features. However, VLP models are not explicitly optimized for fine-grained localization during pre-training. Consequently, as shown in Fig.~\ref{fig:intro}a, raw patch-text scores yield noisy activations, hindering precise semantic object localization, particularly under the weak image-level supervision common in MLR. This imprecise selection limits the quality of derived class-specific features. Therefore, a core challenge is the adaptive refinement of this localization process under weak supervision to obtain higher-quality refined intra-class visual semantic features for improved recognition.

Beyond refining features for individual classes, effectively modeling the complex relationships between classes is crucial for robust scene understanding, particularly in OV-MLR where reasoning about unseen objects often depends on context provided by known entities. While modeling class dependencies is beneficial even in traditional MLR~\cite{ssgrl, ChenLCHW22pami, scenegnn}, approaches relying on co-occurrence statistics mined from the training set are fundamentally incompatible with the open-vocabulary setting, as illustrated in Fig.~\ref{fig:intro}(b). Such statistical information is simply unavailable for novel classes, preventing the generalization of these learned correlations. Existing VLP-based OV-MLR methods also typically lack explicit mechanisms to incorporate and adaptively leverage diverse, generalizable relational knowledge. This underscores the second core challenge: enabling adaptive transfer of information based on rich inter-class relationships that hold true for both seen and unseen categories, moving beyond dataset-specific statistics.

To address these challenges of adaptive refinement and adaptive transfer, we introduce the \textbf{Dual Adaptive Refinement Transfer (DART)} framework. DART enhances a frozen VLP backbone by concurrently tackling both issues through two synergistic modules. For intra-class adaptive refinement, DART incorporates a lightweight \textbf{Adaptive Refinement Module (ARM)} combined with a theoretically grounded \textbf{Weakly Supervised Patch Selecting (WPS) loss}. This combination learns to adaptively refine patch features and localize objects using only image-level annotations. For inter-class adaptive transfer, an \textbf{Adaptive Transfer Module (ATM)} leverages an LLM-mined Class Relationship Graph (CRG) to share relational knowledge. Crucially, the ATM's transfer process builds upon the clean, localized features from the ARM, ensuring relational reasoning is grounded in accurate visual evidence, which in turn provides contextual cues that help refine localization in a positive feedback loop.

Our main contributions are summarized as follows:

1. We propose the Dual Adaptive Refinement Transfer (DART) framework, enhancing frozen VLP models for OV-MLR through synergistic adaptive refinement for localization and adaptive transfer for relational reasoning.

2. We introduce a lightweight Adaptive Refinement Module (ARM) coupled with a theoretically grounded Weakly Supervised Patch Selecting (WPS) loss, enabling adaptive patch feature refinement and discriminative localization to provide a high-quality foundation for relational transfer.

3. We pioneer the use of an LLM-mined Class Relationship Graph (CRG) to guide an Adaptive Transfer Module (ATM), which translates the refined semantic localization into context-aware recognition for both seen and unseen classes by leveraging diverse inter-class relationships.

4. We achieve new state-of-the-art (SOTA) performance on challenging OV-MLR benchmarks, including NUS-WIDE, MS-COCO, and Open Images, which validates the joint optimization of adaptive refinement and transfer.

\begin{figure*}[!t]
    \centering
    \includegraphics[width=\textwidth]{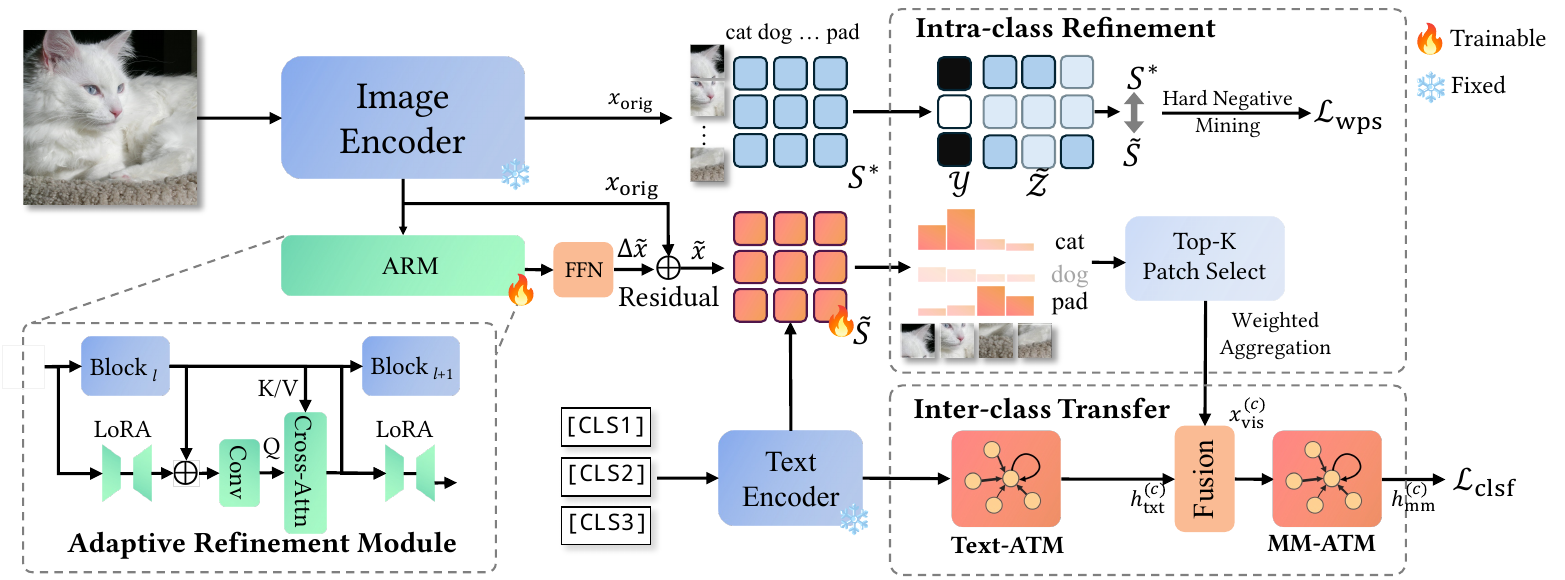}
    \caption{Overall architecture of the DART framework. A frozen VLP backbone processes the input image. The Adaptive Refinement Module (ARM) refines patch features for improved semantic localization, coupled with the WPS loss, yielding class-attentive visual features adaptively. Concurrently, the Adaptive Transfer Module (ATM) leverages an LLM-derived Class Relationship Graph (CRG) to transfer refined information across classes. It first enhances initial text embeddings (Text-ATM) and then processes fused visual-text features (MM-ATM) to produce final relationally-aware representations for prediction.}
    \Description{}
    \label{fig:architecture}
    \vspace{1ex}
\end{figure*}

\section{Related Work}

\noindent\textbf{Traditional MLR.}~
Traditional multi-label methods often consider visual local features and label correlations. For local information, different regions of an image are typically evaluated based on their contribution to the target categories \citep{wei2015hcp, wang2017multi, local3-2021-tip, ChenPWXLL22pami}. For label correlations, semantic interactions between classes are achieved using graphs or other methods, as seen in \cite{ssgrl, chen2019multi, scenegnn, sim-mlr}, which leverage co-occurrence or label similarity information to enable inter-category interactions. However, in the task of multi-label zero-shot learning, where unseen classes need to be recognized, an intuitive approach \citep{hierse, lee} is to establish a connection between unseen and known classes by utilizing pretrained word embeddings such as GloVe \citep{glove} and lexical databases like WordNet. Recent studies, such as LESA \citep{lesa} and BiAM \citep{biam}, based on Glove, capture both regional and global features for better multi-object recognition. While these methods facilitate information transfer between classes through language modalities and have shown some success, they struggles to address the challenges posed by open-vocabulary tasks.

\noindent\textbf{Open-vocabulary MLR.}~
In recent years, with the development of VLP models~\citep{li2019visualbert, li2021align, bao2022vlmo}, open-vocabulary classification has emerged as an alternative to zero-shot prediction, achieving significant progress. Different OV settings in various application tasks, such as detection~\citep{ov-od-cvpr21, ov-od-eccv22, ov-od-cvpr22-2}, segmentation \citep{ov-sg-cvpr-22, ov-sg-eccv-22-2} and scene understanding~\cite{ov-scene-cvpr23, ov-scene-cvpr23-2}, have also been extensively explored. Leveraging billions of image-text pairs as training data, models like CLIP~\citep{clip} and ALIGN~\citep{jia2021scaling} have achieved impressive performance in single-label few-/zero-shot classification tasks \cite{lin2024tagclip,WuYZTWL24tmm,WuCLCWL25TIP}. However, these methods are not fully adaptable to OV-MLR because VLP models are pretrained for single-label classification by learning from one image-text pair, making them easily influenced by the image's dominant category. Consequently, recent works have begun exploring the use of VLP models for OV-MLR tasks. MKT~\citep{mkt} proposed a multi-modal knowledge transfer framework within VLP models, along with a dual-stream module for capturing both local and global features. However, MKT relies on image-level supervision, which is suboptimal for guiding precise patch-level feature selection needed for discriminative localization. Additionally, it lacks an explicit mechanism to leverage structured external knowledge for modeling inter-class relationships. Our proposed DART model addresses these gaps through adaptive patch refinement guided by weak supervision and a knowledge-guided graph interactor that explicitly models relational semantics.

\section{Method}

\subsection{Overall Framework}

Our Dual Adaptive Refinement Transfer (DART) model enhances a frozen vision-language pre-training (VLP) backbone for open-vocabulary multi-label recognition (OV-MLR) by concurrently addressing the key challenges of localization and relational reasoning. As depicted in Fig.~\ref{fig:architecture}, DART integrates two synergistic modules designed for distinct adaptive processes: an Adaptive Refinement Module (ARM) performing intra-class adaptive refinement of patch features for improved localization, and an Adaptive Transfer Module (ATM) executing inter-class adaptive transfer of relational information between class representations.

To achieve fine-grained intra-class adaptive refinement without modifying the frozen VLP, the lightweight ARM attaches to the later layers of the VLP's image encoder (e.g., ViT). Operating parasitically, it processes the original patch features $\mathbf{x}_\text{orig}$ to generate adaptive refinements $\Delta \tilde{\mathbf{x}}$ (Sec. 3.2), yielding adapted patch features $\tilde{\mathbf{x}} = \mathbf{x}_\text{orig} + \Delta \tilde{\mathbf{x}}$. These refined features enable more precise patch-class scoring $\tilde{S}(i,c) = \text{sim}(\tilde{x}^i, t_c)$. The ARM's refinement process is guided by the Weakly Supervised Patch Selecting ($\mathcal{L}_\text{wps}$) loss (Sec. 3.3), which facilitates discriminative patch selection using only image-level labels. Based on the resulting scores $\tilde{S}$, relevant patch features are aggregated and combined with the global VLP feature $\bar{x}_\text{orig}$ to form class-attentive visual features $x_\text{vis}^{(c)}$, better capturing the visual evidence for each class.

To enable robust reasoning, particularly for unseen classes where statistical correlations fail, DART incorporates inter-class adaptive transfer. This begins by constructing a Class Relationship Graph (CRG) using diverse relational knowledge mined from an LLM with extensive world knowledge via prompt engineering before training. The Adaptive Transfer Module (ATM) (Sec. 3.4) then leverages this CRG to facilitate adaptive information flow between class representations using Graph Attention Networks (GATs). The ATM operates sequentially:
(a) Linguistic Domain Enhancement: It first applies GATs to the initial VLP text embeddings $t^{(c)}$ over the CRG, producing relationally enriched text features $h_\text{txt}^{(c)}$.
(b) Multi-modal Domain Interaction: The class-attentive visual features $x_\text{vis}^{(c)}$ are then fused with $h_\text{txt}^{(c)}$, and the ATM again applies GAT-based adaptive transfer on these fused features within the CRG structure, yielding the final relationally-aware multi-modal class representations $h_\text{mm}^{(c)}$.

Final predictions are obtained by computing the similarity between the enhanced multi-modal features $h_\text{mm}^{(c)}$ and the initial text embeddings $t_c$: $\hat y=\text{sim}(h_\text{mm}^{(c)}, t_c)$. Following prior work~\citep{mkt}, we use a ranking loss as the classification objective $\mathcal{L}_\text{clsf}$. To ensure the ARM's adaptive refinement does not drastically alter the original VLP feature space, potentially causing catastrophic forgetting, we introduce an L1 penalty on its output refinements: $\mathcal{L}_\text{penalty} = \Vert \Delta \tilde{\mathbf{x}} \Vert_1$. The overall training objective combines these components:
\begin{equation}
    \mathcal{L} = \mathcal{L}_\text{clsf} + \gamma_\text{wps} \cdot \mathcal{L}_\text{wps} + \gamma_\text{penalty} \cdot \mathcal{L}_\text{penalty}\,,
    \label{eq:loss_overall_final}
\end{equation}
where $\gamma_\text{wps}$ and $\gamma_\text{penalty}$ are balancing hyper-parameters.

\subsection{Intra-Class Semantic Refinement}
\label{sec:arm}

Effective intra-class adaptive refinement of patch features ($\mathbf{x}_\text{orig}$) from the frozen VLP backbone is crucial for precise semantic object localization in OV-MLR, especially given the weak image-level supervision common in MLR datasets. 
Standard VLP features often lack the necessary discriminability to pinpoint specific object instances within an image. To address this, we introduce the Adaptive Refinement Module (ARM). This lightweight module operates parasitically alongside the later layers of the frozen image encoder (e.g., ViT). Its goal is to generate input-adaptive, task-specific feature refinements ($\Delta\tilde{\mathbf{x}}$) that enhance localization capabilities for each class, while preserving the VLP's valuable open-vocabulary knowledge encoded in the frozen weights. The ARM processes features through three key operations sequentially within each designated layer $l$.

\noindent\textbf{LoRA-Augmented Attention Adaptation.} First, to efficiently adapt the self-attention mechanism towards task-relevant spatial patterns crucial for refinement, we employ Low-Rank Adaptation (LoRA)~\cite{hu2022lora}. Trainable low-rank matrices are injected into the query ($\mathbf{W}_q$) and output ($\mathbf{W}_\text{out}$) projection layers of the original attention block. This allows the module to adapt attention weights with minimal parameter overhead. The LoRA-adapted features $\mathbf{x}_\text{arm}^\text{lora}$ used internally by the ARM are computed as:
\begin{align}
&\mathbf{Q} = (\mathbf{W}_q + \mathbf{B}_q\mathbf{A}_q)\mathbf{x}_\text{orig}^{l-1}, \quad \mathbf{K} = \mathbf{W}_k\mathbf{x}_\text{orig}^{l-1}, \quad \mathbf{V} = \mathbf{W}_v\mathbf{x}_\text{orig}^{l-1} \\
&\mathbf{x}_\text{arm}^\text{lora} = (\mathbf{W}_\text{out} + \mathbf{B}_\text{out}\mathbf{A}_\text{out})\cdot\text{Attn}(\mathbf{Q},\mathbf{K},\mathbf{V}) + \mathbf{x}_\text{orig}^{l-1}\,, \label{eq:arm_lora_revised_internal} 
\end{align}
where $\mathbf{x}_\text{orig}^{l-1}$ is the output from the previous original VLP block, and $\mathbf{B}, \mathbf{A}$ are the low-rank matrices ($r \ll d$). Note that this adaptation only affects the internal ARM pathway; the main VLP stream remains unchanged.

\noindent\textbf{Local Context Encoding.} Next, to strengthen local spatial inductive biases often missing in standard Vision Transformers and thereby enhance the features for fine-grained localization refinement, we process the LoRA-adapted features. Specifically, $\mathbf{x}_\text{arm}^\text{lora}$ (excluding the [CLS] token) is reshaped into a spatial grid ($H \times W \times d$) and passed through a lightweight depth-wise convolution (DWConv)~\cite{chollet2017dwconv}:
    $\mathbf{x}_\text{arm}^\text{dw} = \text{DWConv}(\mathbf{x}_\text{arm}^\text{lora})$.
This step efficiently models interactions between spatially adjacent patch features, capturing local structural details vital for distinguishing object parts.

\noindent\textbf{Cross-Attention Feature Integration.} Finally, to ensure the learned local refinements are contextually grounded and effectively integrated with the robust global semantics from the original VLP features ($\mathbf{x}_\text{orig}^l$), we employ a cross-attention mechanism. The encoded ARM features query the corresponding original VLP block's output features, which serve as keys and values:
    $\mathbf{x}_\text{arm}^l = \text{Attn}(Q=\mathbf{x}_\text{arm}^\text{dw}, K=\mathbf{x}_\text{orig}^l, V=\mathbf{x}_\text{orig}^l)$.
This mechanism dynamically gates and blends the refined local information based on its relevance within the broader context provided by the original VLP features $\mathbf{x}_\text{orig}^l$.

The features $\mathbf{x}_\text{arm}^L$ output by the final ARM layer $L$ encapsulate the accumulated task-specific enhancement aimed at improving localization. These are passed through a FFN to compute the final residual feature $\Delta\tilde{\mathbf{x}} = \text{FFN}(\mathbf{x}_\text{arm}^L)$. This residual is then added element-wise to the corresponding original patch features $\mathbf{x}_\text{orig}^L$ from the VLP backbone to produce the final adaptively refined patch features: $\tilde{\mathbf{x}} = \mathbf{x}_\text{orig}^L + \Delta\tilde{\mathbf{x}}$. These adapted features $\tilde{\mathbf{x}}$ thus synergize the VLP's general semantic understanding with enhanced intra-class localization cues, making them suitable for the fine-grained patch scoring and selection detailed in Section~\ref{sec:wps}. Crucially, the ARM design fully preserves the frozen VLP backbone, introducing minimal computational overhead and only a small number of trainable parameters. This facilitates efficient adaptation for improved localization in OV-MLR without risking catastrophic forgetting of the VLP's generalization capabilities.

\begin{figure}[!t]
    \centering
    \includegraphics[width=0.9\linewidth]{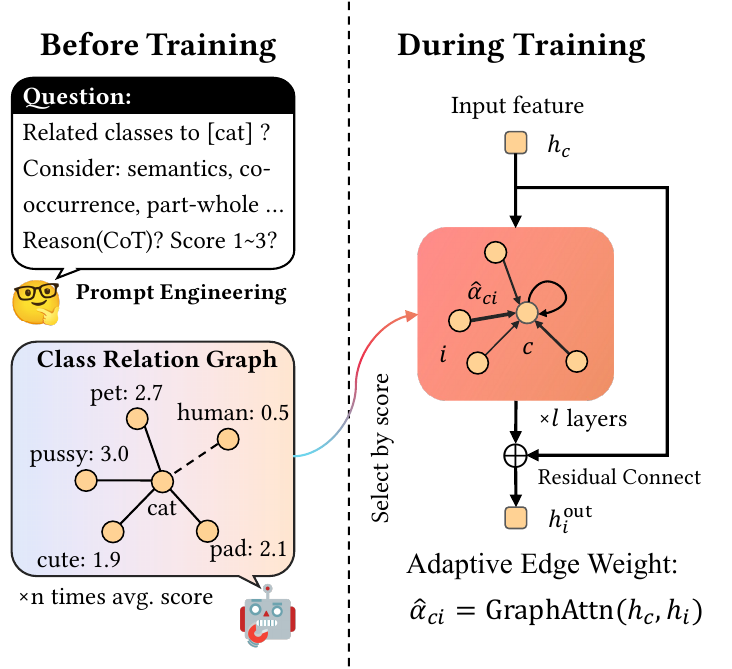}
    \caption{Overview of CRG construction and ATM process.}
    \Description{Overview of CRG construction and ATM process}
    \label{fig:kag}
    \vspace{0.2ex}
\end{figure}

\subsection{Weakly Supervised Patch Selection}
\label{sec:wps}
The ARM provides adapted patch features $\tilde{\mathbf{x}} = \{\tilde{x}^1, \dots, \tilde{x}^{N_p}\}$ (Sec. 3.2). To train the ARM effectively using only image-level labels $\mathcal{Y}=\{y_c\}_{c=1,\dots,N_c}$, we introduce the weakly supervised patch selection (WPS) loss. This loss aims to guide the ARM to produce features that help discriminate class-relevant patches. We define the predicted patch-class score as $\tilde{S}_{i,c} = \text{sim}(\tilde{x}^i, t_c)$ and the score based on original VLP features as $S^*_{i,c} = \text{sim}(x^i_\text{orig}, t_c)$.

We frame this in the Multiple Instance Learning (MIL) paradigm~\citep{mil-review-2010}, where an image is a bag of patches $\mathcal{X}=\{x^i\}_{i=1,\dots,N_p}$. Let $z_{i,c} \in \{0, 1\}$ be the unobserved latent variable indicating if patch $i$ truly corresponds to class $c$. Standard MIL assumptions connect $z_{i,c}$ to the image-level label $y_c$. We need to estimate the posterior probability $\hat{z}_{i,c} = P(z_{i,c}=1 | y_c, \mathcal{X}; \theta)$, representing the responsibility of patch $i$ for class $c$. For positive classes ($y_c=1$), we approximate this responsibility using a Softmax function over the predicted patch scores within the bag: $\hat{z}_{i,c} = \text{Softmax}_j(\tau \cdot \tilde{S}_{i,c})$, where $\tau$ is a learnable temperature parameter. For negative classes ($y_c=0$), the responsibility is zero $\hat{z}_{i,c} = 0$. Note that using Softmax is a common simplification for tractability, rather than a direct derivation from Bayes' rule under MIL assumptions; further justification is provided in the appendix.

Following the Expectation-Maximization (EM) framework, the estimated responsibilities $\hat{z}_{i,c}$ (E-step) guide the parameter updates (M-step). The standard M-step objective minimizes the negative expected complete-data log-likelihood, often realized as a Binary Cross-Entropy (BCE) loss between the predicted scores and the estimated responsibilities: $\sum_{i,c} \text{BCE}(\sigma(\tilde{S}_{i,c}), \hat{z}_{i,c})$.

We refine this objective for better performance and efficiency. First, to stabilize training and leverage the VLP's initial knowledge, we smooth the estimated responsibility for positive classes ($y_c=1$) by interpolating with a prior responsibility $\hat{z}^*_{i,c} = \text{Softmax}_j(\tau \cdot S^*_{j,c})$ derived from the original scores: $\hat{z}'_{i,c} = \lambda \hat{z}^*_{i,c} + (1-\lambda) \hat{z}_{i,c}$. The weight $\lambda$ can be annealed during training. Second, for positive classes ($c \in \mathcal{P}_b = \{c \mid y_c = 1\}$), we simplify the BCE loss term $-[\hat{z}'_{i,c} \log \sigma(\tilde{S}_{i,c}) + (1-\hat{z}'_{i,c}) \log(1-\sigma(\tilde{S}_{i,c}))]$. Since $\hat{z}'_{i,c}$ already encodes patch importance and negative patches receive ample supervision from the negative class loss term, we retain only the first part, focusing on maximizing the evidence for positive patches weighted by their responsibility: $-\hat{z}'_{i,c} \log \sigma(\tilde{S}_{i,c})$. Third, for negative classes ($c \in \mathcal{N}_b = \{c \mid y_c = 0\}$), where $\hat{z}_{i,c}=0$, the BCE simplifies to $-\log(1-\sigma(\tilde{S}_{i,c}))$. To avoid computation over numerous easy negatives, we apply Hard Negative Mining (HNM). We select only the top-$K$ patches with the highest predicted scores $\tilde{S}_{j,c}$ (using stop-gradient, $\text{sg}[\cdot]$) for each negative class, denoted by the index set $\mathcal{I}_{b,c}^{\text{hard}} = \text{TopK}_j (\text{sg}[\tilde{S}_{j,c}])$.

Combining these refinements yields the final WPS loss:
\begin{align}
\mathcal{L}_\text{WPS} = 
\sum_{b \in \mathcal{B}} \Big(
    &\underbrace{
        -\sum_{c \in \mathcal{P}_b} \sum_{i=1}^{N_p} 
        \big(\lambda \hat{z}^*_{i,c} + (1-\lambda) \hat{z}_{i,c}\big) 
        \log \sigma(\tilde{S}_{i,c})
    }_{\text{Weighted Positive Loss}} \label{eq:wps_final_revised} \\
    &\underbrace{
        -\sum_{c \in \mathcal{N}_b} 
        \sum_{j \in \mathcal{I}_{b,c}^{\text{hard}}} 
        \log(\sigma(-\tilde{S}_{j,c}))
    }_{\text{Hard Negative Loss}} \Big). \notag
\end{align}
Interestingly, this formulation connects to recent contrastive methods. The positive term, $-\hat{z}'_{i,c} \log \sigma(\tilde{S}_{i,c})$, resembles the sigmoid loss used in SigLIP~\cite{zhai2023siglip}, which is typically $-\log \sigma(\tilde{S}_{i,c})$. Our loss introduces the dynamically computed responsibility $\hat{z}'_{i,c}$ as an instance weight. Since $\hat{z}'_{i,c}$ is influenced by the model's own predictions (via $\hat{z}_{i,c}$ derived from Softmax over $\tilde{S}$), the loss effectively functions as an instance-weighted contrastive objective. It dynamically focuses the optimization on the patch-class pairs currently considered most salient by the model, providing theoretical justification for its effectiveness in learning discriminative patch representations under weak supervision. Minimizing $\mathcal{L}_\text{WPS}$ thus trains the ARM module to generate adapted features $\tilde{\mathbf{x}}$ conducive to accurate patch selection using only image-level labels.

\begin{table*}[!t]
\resizebox{\textwidth}{!}{
    \begin{tabular}{cc ccccccc | ccccccc}
        \toprule[1pt]
        \multirow{2}{*}{\textbf{Method}} & \multirow{2}{*}{\textbf{Type}} & 
        \multicolumn{7}{c|}{\textbf{Zero-Shot Learning (ZSL)}} & 
        \multicolumn{7}{c}{\textbf{Generalized Zero-Shot Learning (GZSL)}} \\
         & & P@3 & R@3 & F1@3 & P@5 & R@5 & F1@5 & mAP 
            & P@3 & R@3 & F1@3 & P@5 & R@5 & F1@5 & mAP \\
        \midrule
        Fast0Tag\citep{rahman2018deeptag}& \multirow{6}{*}{ZS}
                & 22.6 & 36.2 & 27.8 & 18.2 & 48.4 & 26.4 & 15.1
                & 18.8 &  8.3 & 11.5 & 15.9 & 11.7 & 13.5 &  3.7 \\
        LESA\citep{lesa}& 
                & 25.7 & 41.1 & 31.6 & 19.7 & 52.5 & 28.7 & 19.4 
                & 23.6 & 10.4 & 14.4 & 19.8 & 14.6 & 16.8 & 5.6 \\
        (ML)$^2$-Enc\citep{liu20232mlp2}& 
                &    - &    - & 32.8 &    - &    - & 32.3 & 29.4 
                &    - &    - & 15.8 &    - &    - & 19.2 & 10.2 \\
        SDL$_\text{M=7}$\citep{ben2021sdl}& 
                & 24.2 & 41.3 & 30.5 & 18.8 & 53.4 & 27.8 & 25.9 
                & 27.7 & 13.9 & 18.5 & 23.0 & 19.3 & 21.0 & 12.1 \\
        BiAM\citep{biam}& 
                & 26.6 & 42.5 & 32.7 & 20.5 & 54.6 & 29.8 & 25.9
                & 25.2 & 11.1 & 15.4 & 21.6 & 15.9 & 18.2 &  9.4 \\
        TGF\citep{ma2023tfg}& 
                & 29.0 & 46.3 & 35.6 & 21.4 & 56.9 & 31.1 & 31.1
                & 33.9 & 14.9 & 20.7 & 29.1 & 21.4 & 24.6 & 15.8 \\
        \midrule
        CLIP\citep{clip}& \multirow{4}{*}{OV}
                & 27.0 & 33.5 & 29.9 & 21.2 & 43.8 & 28.5 & 33.7 
                & 31.4 & 13.7 & 19.0 & 26.0 & 18.9 & 21.9 & 16.5 \\
        TagCLIP\citep{lin2024tagclip}& 
                & 26.4 & 42.1 & 32.4 & 19.3 & 51.5 & 28.1 & \ul{38.5}
                & 24.3 & 10.7 & 14.9 & 19.6 & 14.4 & 16.0 & 13.1 \\
        MKT\citep{mkt}& 
& \ul{27.7} & \ul{44.3} & \ul{34.1} & \ul{21.3} & \ul{57.0} & \ul{31.1} & 37.6
& \ul{35.9} & \ul{16.8} & \ul{22.0} & \ul{29.9} & \ul{22.0} & \ul{25.4} & \ul{18.3} \\
        Ours   & 
& \bf{29.3} & \bf{46.9} & \bf{36.1} & \bf{22.8} & \bf{60.8} & \bf{33.2} & \bf{43.9} 
& \bf{39.0} & \bf{17.2} & \bf{23.8} & \bf{32.2} & \bf{23.6} & \bf{27.3} & \bf{22.2} \\
        \bottomrule[1pt]
    \end{tabular}
}
\caption{Zero-shot (ZS) and open-vocabulary (OV) multi-label recognition (MLR) results (\%) on the NUS-WIDE dataset under ZSL and GZSL settings. Best results are \textbf{bolded}, second-best are \ul{underlined}.}
\label{tab:nus-res}
    \vspace{-3ex}
\end{table*}

\subsection{Inter-Class Knowledge Transfer}
\label{sec:atm}

While the ARM focuses on refining intra-class representations for better localization, effective OV-MLR also necessitates robust inter-class reasoning. Modeling relationships between categories can significantly aid recognition, especially for unseen classes or objects where context derived from known classes is crucial. However, traditional approaches relying on training set co-occurrence statistics are inherently limited in the open-vocabulary setting, as such statistics are unavailable for novel categories. Furthermore, simple semantic similarity captured by VLP embeddings may not encompass the diverse functional, spatial, or hierarchical relationships vital for scene understanding. To address this, we introduce the \textbf{Adaptive Transfer Module (ATM)}, designed to facilitate \textit{adaptive inter-class information transfer} by leveraging structured relational knowledge and graph attention mechanisms.

The foundation of the ATM is a Class Relationship Graph (CRG), $\mathcal{G}=(V, E)$, constructed offline before training. Here, $V$ represents the set of all potential target classes (both seen and unseen). As illustrated in the left panel of Fig.~\ref{fig:kag}, we populate this graph by mining diverse relational knowledge from a Large Language Model (LLM) \cite{gpt4,HuangLZWL25acl,HuangL3CZW05ArXiv}. Using carefully engineered prompts, we query the LLM to identify related classes based on multiple criteria (e.g., semantic similarity, spatial co-location, functional interaction). 
To enhance reliability and mitigate potential LLM inconsistencies, each query prompts for: (1) an explicit justification for the proposed relationship (akin to Chain-of-Thought reasoning), promoting grounded connections; and (2) a numerical relevance score. We perform multiple queries per class and aggregate the results, typically averaging scores for consistently identified relationships. 
Finally, we select the top-N scoring neighbors for each class based on these aggregated scores to form the edges $E$ of the final unweighted graph $\mathcal{G}$. This offline process yields a CRG that encodes rich, generalizable world knowledge about class interdependencies, providing the structure for adaptive transfer.

\noindent\textbf{Adaptive Information Transfer with GAT.}
During training, the ATM utilizes the pre-computed graph $\mathcal{G}$ to enable adaptive knowledge transfer between class representations, as shown in the right panel of Fig.~\ref{fig:kag}. Crucially, instead of treating graph edges as fixed conduits, we employ Graph Attention Networks (GATs)~\cite{gat, gatv2}. GATs dynamically compute attention weights $\hat{\alpha}_{i,j}$ between connected nodes (classes) $i$ and $j$ based on their current feature representations $h_l^{(i)}$ and $h_l^{(j)}$: $\hat{\alpha}_{i,j} = \text{GraphAttn}(h_l^{(i)}, h_l^{(j)})$. This mechanism allows the model to adaptively modulate the information flow from neighboring classes based on the specific context, determining which relationships are most relevant for the current representation update. This context-dependent weighting is key to achieving effective adaptive transfer.

The ATM architecture applies this GAT-based adaptive transfer process sequentially across two domains:
(1)  \textit{Linguistic Domain Enhancement (Text-ATM)} 
takes the initial VLP text embeddings $t_c$ as input ($h_0^{(c)} = t_c$). It performs $l_G$ layers of GAT-based message passing on $\mathcal{G}$ to produce relationally enhanced text embeddings $h_\text{txt}^{(c)} = h_{l_G}^{(c)}$.
(2) \textit{Multi-modal Domain Interaction (MM-ATM)}  
takes the fused visual-text features $x_\text{mm}^{(c)}$ (derived from $x_\text{vis}^{(c)}$ and $h_\text{txt}^{(c)}$ by an FNN: $x_\text{mm}^{(c)}=\text{FNN}([x_\text{vis}^{(c)}\mid h_\text{txt}^{(c)}])$) as input ($h_0^{(c)} = x_\text{mm}^{(c)}$). It applies another $l_G$ layers of GAT interaction on $\mathcal{G}$ to output the final, relationally-aware multi-modal representations $h_\text{mm}^{(c)} = h_{l_G}^{(c)}$.

Within each layer $l$ of the GAT, the representation $h_l^{(c)}$ for class $c$ is updated by aggregating information from its neighbors $\mathcal{N}(c)$ in $\mathcal{G}$, weighted by the learned attention coefficients $\hat{\alpha}_{ci}$, and combined with the previous representation via a residual connection:
\begin{equation}
h_{l+1}^{(c)} = \sum_{j \in \mathcal{N}(c) \cup \{c\}} \hat{\alpha}_{cj} \mathbf{W}_l h_l^{(j)} + h_l^{(c)},
\label{eq:atmgat_final}
\end{equation}
where $\mathbf{W}_l$ is a learnable linear transformation for layer $l$. This adaptive aggregation mechanism enables the ATM to selectively leverage the most pertinent relational knowledge encoded in the LLM-derived graph $\mathcal{G}$ for the specific task context (linguistic or multi-modal). The final outputs, particularly $h_\text{mm}^{(c)}$, are thus enriched representations suitable for accurate classification. (Further implementation details are in the Appendix).

\begin{table}[t!]
    \centering
    \setlength{\belowrulesep}{1pt}
    \setlength{\tabcolsep}{3pt}
    
    \begin{tabular}{c r ccc ccc c}
        \toprule[.75pt]
        \multirow{2}{*}{\textbf{Method}} & 
        \multirow{2}{*}{\textbf{Task}} &
        \multicolumn{3}{c}{\textbf{k=10}} & 
        \multicolumn{3}{c}{\textbf{k=20}} & \multirow{2}{*}{mAP}\\
        &  &    P &    R &   F1 &    P &    R &   F1 & \\
        \midrule
        \multirow{2}{*}{LESA\citep{lesa}} 
            &  ZSL &  0.7 & 25.6 &  1.4 &  0.5 & 37.4 &  1.0 & 41.7 \\ \vspace{-2pt}
            & GZSL & 16.2 & 18.9 & 17.4 & 10.2 & 23.9 & 14.3 & 45.4 \\
        \cmidrule(r){2-9}
        \multirow{2}{*}{ZS-SDL\citep{ben2021sdl}}
            &  ZSL &  6.1 & 47.0 & 10.7 &  4.4 & 68.1 &  8.3 & 62.9 \\ \vspace{-2pt}
            & GZSL & 35.3 & 40.8 & 37.8 & 23.6 & 54.5 & 32.9 & 75.3 \\
        \cmidrule(r){2-9}
        \multirow{2}{*}{BiAM\citep{biam}}
            &  ZSL &  3.9 & 30.7 &  7.0 &  2.7 & 41.9 &  5.5 & 65.6 \\ \vspace{-2pt}
            & GZSL & 13.8 & 15.9 & 14.8 &  9.7 & 22.3 & 14.8 & 81.7 \\
        \cmidrule(r){2-9}
        \multirow{2}{*}{(ML)$^2$-Enc\citep{liu20232mlp2}}
            &  ZSL &    - &    - &  7.5 &    - &    - &  6.5 & 65.7 \\ \vspace{-2pt}
            & GZSL &    - &    - & 27.6 &    - &    - & 24.1 & 79.9 \\
        \cmidrule(r){2-9}
        \multirow{2}{*}{TGF\citep{ma2023tfg}}
            &  ZSL &  4.0 & 31.0 &  7.1 &  3.0 & 47.1 &  5.7 & 63.5 \\ \vspace{-2pt}
            & GZSL & 36.9 & 42.7 & 39.6 & 24.2 & 56.0 & 33.8 & 77.6 \\
        \cmidrule(r){2-9}
        \multirow{2}{*}{CLIP\citep{clip}}
            &  ZSL & 10.8 & 84.0 & 19.1 &  5.9 & 92.1 & 11.1 & 66.2 \\ \vspace{-2pt}
            & GZSL & 37.5 & 43.3 & 40.2 & 25.4 & 58.7 & 35.4 & 77.5 \\
        \cmidrule(r){2-9}
        \multirow{2}{*}{MKT\citep{mkt}}
            &  ZSL 
& \ul{11.1} & \bf{86.8} & \ul{19.7} &  \ul{6.1} & \bf{94.7} & \ul{11.4} & \ul{68.1} \\ \vspace{-2pt}
            & GZSL 
& \ul{37.8} & \ul{43.6} & \ul{40.5} & \ul{25.4} & \ul{58.5} & \ul{35.4} & \ul{81.4} \\ 
        \cmidrule{2-9}
        \multirow{2}{*}{Ours} 
            &  ZSL 
& \bf{12.6} & \ul{85.9} & \bf{21.9} &  \bf{6.2} & \ul{93.5} & \bf{11.5} & \bf{70.0} \\ \vspace{-2pt}
            & GZSL 
& \bf{39.7} & \bf{44.1} & \bf{41.8} & \bf{27.1} & \bf{62.0} & \bf{37.7} & \bf{83.5} \\
        \bottomrule[.75pt]
    \end{tabular}
    \caption{Open-Vocabulary multi-label recognition results (\%) on Open Images dataset.}
    \label{tab:open-res}
\end{table}

\section{Experiments}
\subsection{Experiment Setup}
\noindent\textbf{Dataset.}~
We validate the superiority of our model using three widely recognized benchmarks. \textbf{NUS-WIDE} \citep{nuswide} comprises a training set of 161,789 images and a testing set of 107,859 images. Following the LESA~\citep{lesa} setting, we treat 81 human-verified labels as unseen labels, and 925 labels generated user tags as seen labels. \textbf{MS-COCO}~\citep{coco} dataset dataset is divided into training and validation sets with 82,783 and 40,504 images, respectively. We follow SDL~\citep{ben2021sdl} split classes to 48 seen and 17 unseen classes. The \textbf{Open Images (v4)} \citep{openimages-v4} dataset includes 9,011,219 images for training, 41,620 images for validation, and 125,436 images for testing. As per LESA, we designate 7,186 labels in the training set as seen classes, and the 400 most frequent labels as unseen classes.

\begin{table}[t]
    \centering
    \begin{tabular}{c ccc | ccc}
        \toprule[1pt]
        \multirow{2}{*}{\textbf{Method}} &
        \multicolumn{3}{c|}{\textbf{ZSL}} & 
        \multicolumn{3}{c}{\textbf{GZSL}} \\
                 &    P &    R &   F1 &    P &    R &   F1 \\
        \midrule
        Fast0Tag\citep{zhang2016fasttag}
            & 24.7 & 61.4 & 25.3 & 38.5 & 46.5 & 42.1 \\
        Deep0Tag\citep{rahman2018deeptag}
            & 26.5 & 65.9 & 37.8 & 43.2 & 52.2 & 47.3 \\
        SDL$_\text{M=2}$\citep{ben2021sdl}
            & 26.3 & 65.3 & 37.5 & 59.0 & 60.8 & 59.9 \\
        CLIP\citep{clip}
            & 32.1 & 83.3 & 46.4 & 45.4 & 50.2 & 47.7 \\
        MKT\citep{mkt}
            & \ul{34.3} & \ul{84.5} & \ul{48.9} & \ul{68.3} & \ul{62.0} & \ul{65.0}\\
        Ours
            & \bf{36.6} & \bf{91.0} & \bf{52.2} & \bf{71.2} & \bf{64.7} & \bf{67.8} \\
        \bottomrule[1pt]
    \end{tabular}
    \caption{Open-Vocabulary multi-label recognition results (\%) on MS-COCO dataset.}
    \label{tab:coco-res}
    \vspace{-3ex}
\end{table}

\begin{table}[t]
\setlength{\belowrulesep}{1pt}
\setlength{\tabcolsep}{3pt}
\centering
\begin{tabular}{lcccc|ccc}
\toprule[.75pt]
\multirow{2}{*}{\textbf{Method}} & \multirow{2}{*}{\textbf{Param.}} & 
\multicolumn{3}{c|}{\textbf{ZSL}} & \multicolumn{3}{c}{\textbf{GZSL}} \\
& & P & R & F1 & P & R & F1\\
\midrule
LLaVA1.5
    & 7B    & 32.6 & 20.3 & 25.1 & 43.8 &  8.4 & 14.1 \\
LLaVA1.6
    & 7B    & 35.6 & 17.3 & 23.3 & 48.0 &  7.4 & 12.8 \\
LLaMA3.2V
    & 11B   & 28.8 & \ul{55.5} & 37.9 & 42.1 & \ul{25.6} & 31.8 \\ 
Qwen2.5VL
    & 7B    & \bf{91.7} & 51.5 & \ul{66.0} & \bf{95.1} & 23.0 & \ul{37.0} \\
Ours$^*$
    & 100M   & \ul{81.3} & \bf{62.4} & \bf{70.6} & \ul{78.5} & \bf{37.5} & \bf{50.8} \\
\bottomrule[.75pt]
\end{tabular}%
\caption{Comparison with MLLMs on the subset of NUS-WIDE. ($*$: our model predicts labels at a 0.5 threshold.)}
\label{tab:mllm-res}
    \vspace{-3ex}
\end{table}

\noindent\textbf{Metrics.}~
Following previous works, we adapt the precision (P), recall (R) and F1 score (F1) per sample to evaluate models. Additionally, we also introduce the metric of mean average precision (mAP) over all categories for NUS-WIDE and Open Images datasets.
We evaluate under two standard settings: Zero-Shot Learning (ZSL), where performance is measured only on unseen classes not encountered during training, and Generalized Zero-Shot Learning (GZSL), where evaluation considers both seen and unseen classes simultaneously.
GZSL is significantly more challenging as it demands balanced recognition across both sets. In the standard splits we follow~\citep{lesa, ben2021sdl}, seen classes are often more numerous (925 seen v.s. 81 unseen classes for NUS-WIDE) and may contain rarer or less distinct examples compared to the selected unseen classes. Consequently, as observed across various methods, GZSL performance metrics are typically lower than ZSL metrics due to the difficulty of generalizing without bias towards the potentially more frequent or distinct unseen classes encountered during evaluation.

\begin{figure*}[t!]
    \centering
    \begin{subfigure}[b]{0.247\textwidth}
        \includegraphics[width=\linewidth]{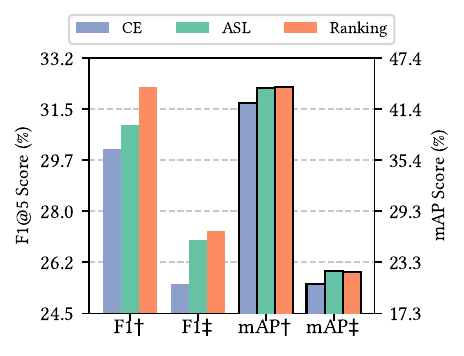}
        \subcaption{Comparison with $\mathcal{L}_\text{clsf}$}
        \label{fig:a_loss_type}
    \end{subfigure}
    \begin{subfigure}[b]{0.247\textwidth}
        \includegraphics[width=\linewidth]{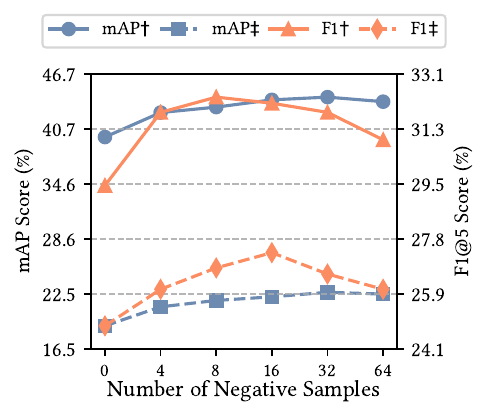}
        \subcaption{Effect of negative patch number}
        \label{fig:b_neg}
    \end{subfigure}
    \begin{subfigure}[b]{0.247\textwidth}
        \includegraphics[width=\linewidth]{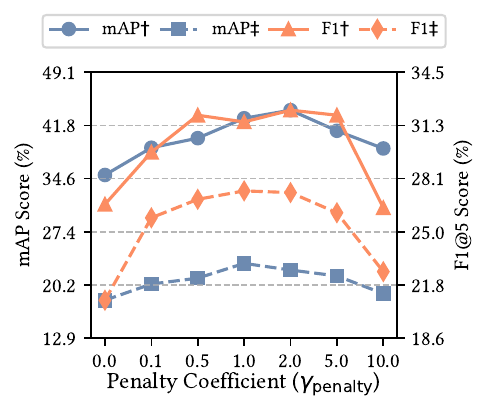}
        \subcaption{Effect of $\gamma_\text{penalty}$}
        \label{fig:c_penalty}
    \end{subfigure}
    \begin{subfigure}[b]{0.247\textwidth}
        \includegraphics[width=\linewidth]{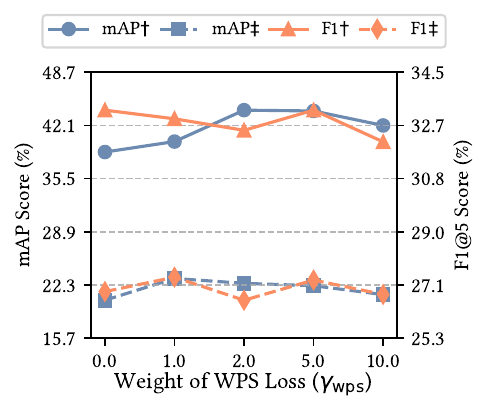}
        \subcaption{Effect of $\gamma_\text{wps}$}
        \label{fig:d_wps_weight}
    \end{subfigure}\\
    \vspace{-2ex}    
    \caption{Ablation experiments on loss components. ($\dagger$: ZSL result; $\ddagger$: GZSL result)}
    \label{fig:loss}
    \vspace{-2ex}
\end{figure*}

\subsection{Comparisons with State-of-the-art Methods}
\noindent\textbf{Comparison with Prior Methods.} Our method consistently sets new SOTA performance across all datasets. As shown in Tab.~\ref{tab:nus-res}, on NUS-WIDE, DART achieves substantial gains over the previous SOTA. Specifically, under the ZSL setting, DART boosts mAP by a significant 14.0\% and F1 score (K=3) by up to 6.8\%. For GZSL, the improvements are even more pronounced, with mAP increasing by 21.3\% and F1 by up to 8.2\%. On the large-scale Open Images dataset shown in Tab.~\ref{tab:open-res}, DART continues to show significant advantages, improving ZSL F1 score by a large margin of 11.2\% (at K=10) and GZSL F1 by 6.5\% (at K=20). Similarly, on MS-COCO, DART surpasses the previous best F1 performance by 6.7\% for ZSL and 4.3\% for GZSL (K=3) in Tab.~\ref{tab:coco-res}. These consistent and significant improvements across diverse datasets firmly establish the effectiveness and superiority of our proposed model.

\newcommand{\ys}{\ding{51}}
\newcommand{\no}{}

\begin{table}[t!]
    \centering
    \setlength{\tabcolsep}{1mm}
    \setlength{\belowrulesep}{1pt}
    \begin{tabular}{ccc ccc ccc}
        \toprule[.75pt]
        \multicolumn{3}{c}{\textbf{Method}} & 
        \multicolumn{3}{c}{\textbf{ZSL}} & 
        \multicolumn{3}{c}{\textbf{GZSL}} \\
        ARM & WPS & ATM & F1@3 & F1@5 & mAP & F1@3 & F1@5 & mAP \\
        \midrule
        \no & \no & \no & 29.9 & 28.5 & 33.7 & 19.0 & 21.9 & 16.5 \\
        \ys & \no & \no & 32.3 & 28.8 & 37.4 & 20.8 & 23.4 & 19.5 \\
        \ys & \no & \ys & \ul{35.4} & \bf{33.2} &     38.8  & \ul{22.9} & \ul{26.9} &     20.4 \\
        \ys & \ys & \no &     34.0  &     31.9  & \bf{44.2} &     22.7  &     25.0  & \ul{22.1} \\
        \ys & \ys & \ys & \bf{36.1} & \bf{33.2} & \ul{43.9} & \bf{23.8} & \bf{27.3} & \bf{22.2} \\
        \bottomrule[.75pt]
    \end{tabular}
    \caption{Ablation result (\%) of proposed modules.}
    \label{tab:module_ablation}
    \vspace{-3ex}
\end{table}

\noindent\textbf{Comparison with Multi-Modal LLMs (MLLMs).}
Beyond comparing with methods specifically designed for MLR, we also evaluate DART against the distinct paradigm of MLLMs, which approach recognition tasks generatively.
We also compare DART with recent MLLMs on a NUS-WIDE subset, because of given the large model sizes (7B-11B) and substantial inference costs of MLLMs like LLaVA-1.5 (7B)~\citep{llava}, LLaVA-1.6 (7B)~\citep{llava_1_6}, Llama3.2V (11B)~\citep{llama3}, and Qwen2.5-VL (7B)~\citep{qwen2,qwen2_5}. As shown in Tab.~\ref{tab:mllm-res}, DART, with only \textbf{123}M parameters, demonstrates superior F1 performance. Notably, it surpasses the strongest MLLM, Qwen2.5-VL, by 4.6 points in ZSL F1 and a significant 13.8 points in GZSL F1. Analyzing precision reveals a key weakness in some MLLMs: LLaVA and Llama3.2V exhibit severe hallucination, frequently predicting labels outside of the defined seen/unseen vocabulary, significantly degrading their precision scores. While Qwen2.5-VL achieves slightly higher precision, DART's recall significantly surpasses all compared MLLMs, suggesting better coverage beyond just the high-confidence or major categories often favored by MLLMs. These results underscore that DART provides a highly effective and well-balanced solution for the OV-MLR task, achieving superior F1 scores with lower computational overhead compared to current MLLMs.

\begin{table}[t!]
    \setlength{\belowrulesep}{1pt}
    \centering
    \begin{tabular}{l ccc ccc}
        \toprule[.75pt]
        \multirow{2}{*}{\bf{Relation}} & \multicolumn{3}{c}{\bf{ZSL}} & \multicolumn{3}{c}{\bf{GSL}} \\
         & F1@3 & F1@5 & mAP & F1@3 & F1@5 & mAP\\
         \midrule
         Rand.  & 30.2 & 27.8 & 33.4 & 21.3 & 24.3 & 18.1 \\
         Sim.   & 34.2 & 32.1 & 43.6 & 22.5 & 25.7 & 21.1 \\
         LLM+Rand.& 35.8 & 32.9 & 43.4 & 23.6 & 27.0 & 22.3 \\
         LLM    & 36.1 & 33.2 & 43.9 & 23.8 & 27.3 & 22.2 \\
         \bottomrule[.75pt]
    \end{tabular}
    \caption{Ablation result (\%) of CRG. (Rand.: random relations; Sim.: similarity from text embedding; LLM.: mining from LLM; LLM+Rand.: random replace LLM CRG with noise.)}
    \label{tab:crg}
    \vspace{-3ex}
\end{table}

\begin{figure*}[t!]
    \centering
    
    \begin{minipage}{0.48\textwidth}
        \centering
        \includegraphics[width=\linewidth]{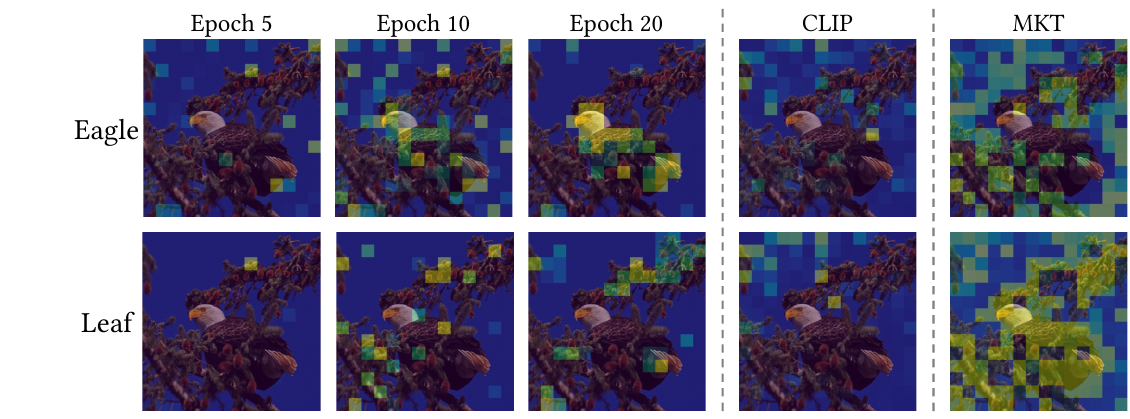}
    \end{minipage}
    \begin{minipage}{0.48\textwidth}
        \centering
        \includegraphics[width=\linewidth]{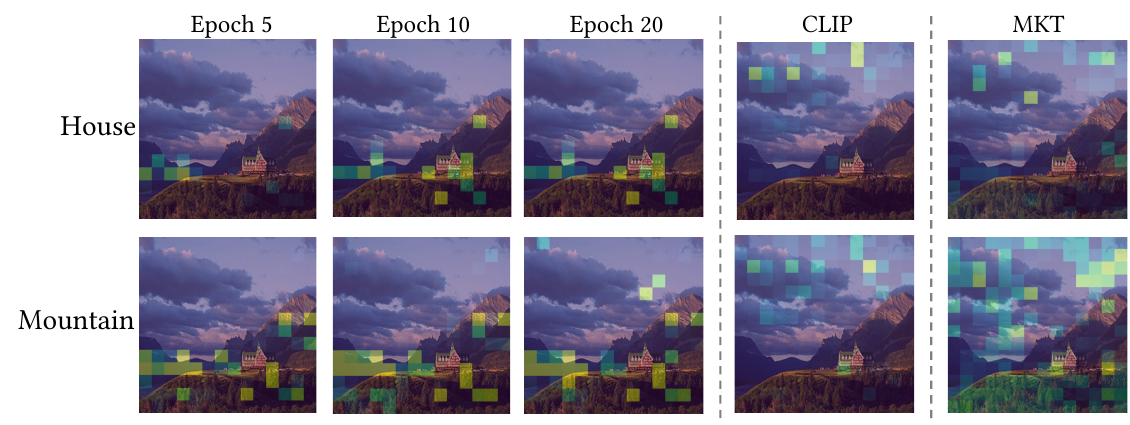}
    \end{minipage}
    \vspace{-2ex}
    \caption{Comparison of patch scores across different models and our approach at various training stages (best viewed in zoom).}
    \label{fig:patch_score}
    \vspace{-1ex}
\end{figure*}

\begin{figure*}[!t]
    \centering
    \includegraphics[width=0.9\textwidth]{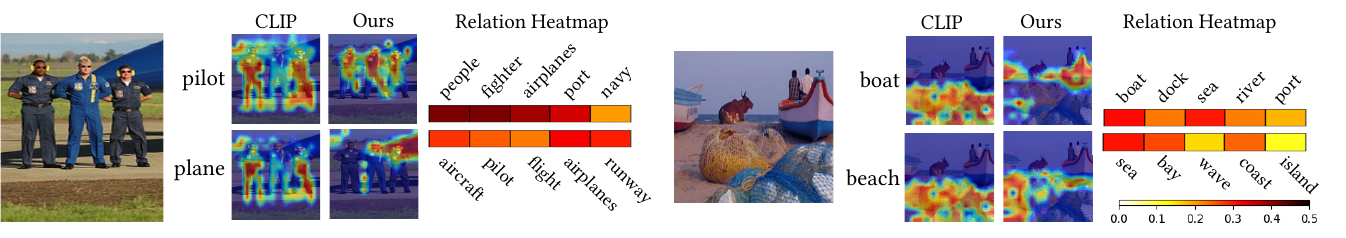}
    \vspace{-2ex}
    \caption{Visualization of the relational classes weights in MM-ATM and the Class Activation Map.}
    \label{fig:relation}
    \vspace{-1ex}
\end{figure*}

\subsection{Ablation Studies}

\noindent\textbf{Ablation on Proposed Modules.}
We evaluate the contribution of each proposed component in Tab.~\ref{tab:module_ablation}. The ARM alone significantly increases CLIP baseline performance (+11. 0\% ZSL mAP, +18. 2\% GZSL mAP). Adding the ATM leverages inter-class relationships, enhancing the prediction of the complete label set per image and thus boosting F1 scores (+15.3\% ZSL F1@5, +18.8\% GZSL F1@5). Alternatively, incorporating the WPS loss sharpens the model's focus on discriminative class-specific visual evidence, increasing mAP (+18.2\% ZSL mAP, +13.3\% GZSL mAP). These distinct improvements highlight the complementary functions of ATM (improving comprehensive image understanding, reflected in F1) and WPS (enhancing class-specific localization accuracy, reflected in mAP). The full DART model integrating ARM, WPS, and ATM achieves the best overall performance, confirming the synergistic benefits of jointly optimizing these crucial aspects.

\noindent\textbf{Ablation on Loss.} 
We performed ablation studies on the loss components, summarized in Fig.~\ref{fig:loss}. First, evaluating alternative classification losses revealed that a ranking loss yielded superior results compared to Cross-Entropy and ASL, likely due to better synergy with CLIP's metric learning pretrain objective (Fig.~\ref{fig:a_loss_type}). Analyzing the hard negative mining in the WPS loss showed its critical importance; performance plummeted without negative examples, yet excessive negatives were counterproductive (Fig.~\ref{fig:b_neg}). We examined the impact of regularizing the residual features of the ARM module. Removing this regularization ($\gamma_\text{penalty}=0$) caused catastrophic forgetting of VLP knowledge, while excessive penalization hampered task-specific adaptation, demonstrating the need for careful balance (Fig.~\ref{fig:c_penalty}). Adjusting the overall weight of the WPS loss indicated that while increasing its influence could boost mAP by enhancing localization, overly high weights tended to slightly decrease the F1 score, highlighting a trade-off between localization guidance and overall classification accuracy (Fig.~\ref{fig:d_wps_weight}).

\noindent\textbf{Effect of Class Relation Graph.}
We further investigate the importance of the knowledge source for the CRG and the robustness of our ATM on Tab.~\ref{tab:crg}. Replacing the LLM-mined CRG with a randomly generated graph leads to a significant performance drop (e.g., -23.9\% mAP for ZSL and -18.5\% mAP for GZSL compared to using the LLM CRG), underscoring the necessity of meaningful relational information. When using a CRG constructed from CLIP text embedding similarities, we observe a slight improvement over the baseline without ATM (+0.6\% F1@3 for ZSL, +2.8\% F1@3 for GZSL), indicating that semantic similarity provides some useful relational cues. However, this similarity-based CRG underperforms the LLM-mined version, particularly in mAP (-11.6\% for ZSL, -8.1\% for GZSL), likely because basic similarity is already implicitly leveraged by the base VLP model, limiting the additional benefit for precise class discrimination. The LLM-mined CRG, capturing more diverse relationships, proves more effective. Furthermore, even when randomly replacing a portion of nodes in the LLM-mined CRG, the performance degradation is minimal. This demonstrates the robustness of our ATM Interactor, whose adaptive graph attention mechanism can effectively mitigate noise and focus on reliable relational information.

\subsection{Qualitative Analysis}

\noindent\textbf{Visualization of Learned Patch Selection.}
As depicted in Fig.~\ref{fig:patch_score}, we visualize the evolution of patch scores learned via our WPS loss. Initially, during early training epochs, the patch selection appears relatively sparse and diffuse. However, as training progresses, the model learns to assign higher scores to patches corresponding to target objects, resulting in selections that accurately highlight the relevant regions. While we observe false positives activation in background patches, the final learned patch selections demonstrate considerably more precise localization compared to the raw patch-text similarities from CLIP or MKT.

\noindent\textbf{Visualization of Category Relationships and Class Activation Map.}
Fig.~\ref{fig:relation} presents Grad-CAM~\citep{gradcam} visualizations and inter-class relationship heatmap weights of MM-ATM. The activation maps clearly show that DART achieves more precise localization of object regions compared to baseline CLIP. Furthermore, the heatmap illustrates the ATM's capability to adaptively transfer information, revealing that the top-5 correlation coefficients often correspond to categories closely related to those present in the image, thereby enhancing recognition.

\section{Conclusion}

We have introduced the Dual Adaptive Refinement Transfer (DART) framework to enhance Open-Vocabulary Multi-Label Recognition (OV-MLR) by achieving precise semantic localization under weak supervision adaptively and addressing the challenge of effectively leveraging structured relational knowledge beyond basic semantics, especially for unseen classes. DART synergistically integrates two novel adaptive modules with a frozen VLP backbone: an Adaptive Refinement Module (ARM) coupled with a Weakly Supervised Patch Selecting (WPS) loss for improved intra-class localization under weak supervision adaptively, and an Adaptive Transfer Module (ATM) leveraging an LLM-derived Class Relationship Graph (CRG) for adaptive inter-class knowledge transfer. DART is the first framework, to our knowledge, to jointly optimize these adaptive refinement and transfer mechanisms, particularly incorporating external LLM knowledge for OV-MLR. Extensive experiments confirm that DART achieves state-of-the-art performance on major benchmarks, validating the efficacy of our proposed approach.

\begin{acks}
This work was supported by National Natural Science Foundation of China (NSFC) under Grant No. 62272494 and 62325605, Guangdong Basic and Applied Basic Research Foundation under Grant No. 2023A1515012845 and 2023A1515011374, and Guangdong Province Key Laboratory of Information Security Technology.
\end{acks}

\bibliographystyle{ACM-Reference-Format}
\bibliography{main}

\balance



\clearpage

\title[Supplementary Material]{Supplementary Material for DART: Dual Adaptive Refinement Transfer for Open-Vocabulary Multi-Label Recognition\\~}
\subtitle{Haijing Liu, Tao Pu, Hefeng Wu, Keze Wang, Liang Lin\\\vspace{0.8ex}Sun Yat-sen University\\\vspace{2ex}~}

\maketitlesupplementary

\appendix

\section{More Implementation Details}
\label{sec:appendix_implementation}

This section provides additional details regarding the experimental setup and hyperparameters used for training our DART framework, complementing the information presented in the main paper.

\noindent\textbf{Training Setup}.
We trained our models using the AdamW optimizer~\citep{loshchilov2017decoupled}. The learning rate was managed using a cosine annealing schedule, starting from an initial value of 1e-4 and gradually decaying to 1e-5 over the course of training. To stabilize training in the initial phase, we employed a linear learning rate warm-up strategy for the first 2 epochs. All experiments were conducted with a batch size of 256. The total number of training epochs was set based on the dataset: 20 epochs for MS-COCO and NUS-WIDE, and 10 epochs for the larger Open Images dataset.

\noindent\textbf{Backbone Architecture}.
As mentioned in the main text, our approach builds upon a frozen Vision-Language Pre-training (VLP) model. Specifically, we utilized the publicly available pre-trained CLIP model~\citep{clip} with the ViT-B/16 architecture as the vision transformer backbone for extracting initial image features. The weights of this backbone were kept frozen throughout training.

\noindent\textbf{Adaptive Refinement Module (ARM)}.
The ARM module incorporates Low-Rank Adaptation (LoRA)~\citep{hu2022lora} to fine-tune specific parameters within the vision backbone's attention layers. We applied LoRA to the Query (Q) and Output (Out) projection matrices. Based on preliminary experiments, we set the LoRA rank $r=32$ and the scaling factor $\alpha=128$. Other components of the ARM, such as the depthwise convolution and cross-attention mechanisms, followed standard implementations as described in the main paper.

\noindent\textbf{Weakly Supervised Patch Selecting (WPS) Loss}.
The temperature parameter $\tau$ used in the WPS contrastive loss (Eq.~3) was not treated as a hyperparameter but was instead initialized directly from the learned temperature value of the pre-trained CLIP model and kept fixed during training. The weight coefficient $\gamma_\text{wps}$ for the WPS loss term and the number of hard negatives $K$ used in the hard negative mining strategy were determined via ablation studies (see Sec. 4.3 in the main paper). Optimal performance was achieved with $\gamma=5$ and $K=16$.

\noindent\textbf{Adaptive Transfer Module (ATM)}.
The Graph Attention Network (GAT) layers within both the Text-ATM (operating on linguistic features) and the MM-ATM (operating on multi-modal features) were implemented using the GATv2 formulation~\citep{gatv2}, known for its potentially improved expressiveness compared to the original GAT. Both the Text-ATM and the MM-ATM modules consisted of 2 GATv2 layers each.

\noindent\textbf{Class Relationship Graph (CRG) Construction}.
The CRG, which provides the structural basis for the ATM, was constructed offline prior to model training. We utilized the `gpt-4o-latest' Large Language Model (LLM) to mine potential relationships between classes. To generate diverse yet relevant connections, we queried the LLM 3 times for each class using a low top-p sampling value of 0.3. The results from these queries were aggregated (e.g., by taking the union of suggested neighbors and selecting the top N based on frequency or LLM confidence, if available) to determine the final neighbors for each class node in the graph. The number of neighbors (N) selected per class was adjusted based on the dataset characteristics:
\begin{itemize}
    \item For NUS-WIDE and MS-COCO datasets, we constructed the CRG by connecting each class to its top N=8 related neighbors identified by the LLM.
    \item For the larger-scale Open Images dataset, we increased the connectivity, setting N=12 neighbors per class.
\end{itemize}
This dataset-dependent configuration allows the graph density to scale appropriately with the label space complexity.

\section{Evaluation Metrics Details}

\subsection{Mean Average Precision}
Following \citet{map-src}, we calculate average precision for each category $c$ as:
\begin{equation}
    \text{AP}_c=\frac{\sum_{n=1}^N\text{Precision}(n, c)\cdot \text{rel}(n,c)}{N_c},\eqno{\text{(B.1)}}\notag
\end{equation}
where \text{Precision}(n, c) is the precision for category $c$ when retrieving $n$ highest-ranked predicted scores and $\text{rel(n, c)}$ is an indicator function that is 1 if the image at rank $n$ contains label $c$ and 0 otherwise. $N_c$ denotes the number of positives for category $c$. Then mAP is computed as:
\begin{equation}
    \text{mAP}=\frac{1}{C}\sum_{c=1}^C\text{AP}_c,\eqno{\text{(B.2)}}\notag
\end{equation}
where $C$ is the number of categories.

\subsection{F1 Score}
Following \citet{f1-src}, we assign $K$ highest-ranked predictions to each image and compare them with the ground truth labels. The mean-per-label precision and mean-per-label recall are calculated as:
\begin{equation}
    P=\frac{\sum_c N_c^\text{TP}}{\sum_c N_c^\text{P}},\quad R=\frac{\sum_c N_c^\text{P}}{\sum_c N_c},\eqno{\text{(B.3)}}\notag
\end{equation}
where $N_c^\text{TP}$ is the number of true positive for category $c$ in top-K prediction and $N_c^{P}$ is the number of positive predictions for category $c$. Therefore, the F1 score is computed as:
\begin{equation}
    \text{F1}=\frac{2PR}{P+R}.\eqno{\text{(B.4)}}\notag
\end{equation}

\section{Theoretical details}

\subsection{Derivation of the Weakly Supervised Patch Selecting (WPS) Loss}
\label{app:wps_derivation}

This appendix provides a detailed derivation of the Weakly Supervised Patch Selecting (WPS) loss presented in Sec.~3.3. The goal is to train the Adaptive Refinement Module (ARM), parameterized by $\theta$, to produce refined patch features $\tilde{\mathbf{x}}$ that allow for effective identification of class-relevant patches using only image-level labels $y = \{y_c\}_{c=1}^C$. The derivation employs the Multiple Instance Learning (MIL) framework and the Expectation-Maximization (EM) algorithm.

\subsubsection{Notation Recap}
\begin{itemize}
    \item $b$: An image (bag) from the training batch $\mathcal{B}$.
    \item $i$: Index for patches within an image, $i \in \{1, ..., N_b\}$.
    \item $c$: Index for classes.
    \item $y_c \in \{0, 1\}$: Image-level label for class $c$.
    \item $z_{i,c} \in \{0, 1\}$: Latent variable indicating if patch $i$ corresponds to class $c$.
    \item $\tilde{S}_{i,c}$: Predicted patch-class score using ARM features $\tilde{x}^i(\theta)$ and class embedding $t^c$, calculated as $\text{sim}(\tilde{x}^i(\theta), t^c)$ potentially scaled.
    \item $S^*_{i,c}$: Original patch-class score using frozen VLP features $x^i_{\text{orig}}$ and $t^c$, calculated as $\text{sim}(x^i_{\text{orig}}, t^c)$ potentially scaled.
    \item $\hat{z}_{i,c}$: Model-predicted responsibility of patch $i$ for class $c$, derived from $\tilde{S}_{i,c}$.
    \item $\hat{z}^*_{i,c}$: Prior responsibility of patch $i$ for class $c$, derived from $S^*_{i,c}$.
    \item $\hat{z}'_{i,c}$: Final smoothed responsibility used in the M-step, combining $\hat{z}_{i,c}$ and $\hat{z}^*_{i,c}$.
    \item $\mathcal{P}_b = \{c | y_c = 1\}$: Set of positive classes for image $b$.
    \item $\mathcal{N}_b = \{c | y_c = 0\}$: Set of negative classes for image $b$.
    \item $\sigma(\cdot)$: Sigmoid function, $\sigma(s) = 1 / (1 + \exp(-s))$.
    \item $\theta$: Parameters of the ARM module being learned.
\end{itemize}

\subsubsection{Multiple Instance Learning (MIL) Assumptions}
The connection between image labels $y_c$ and latent patch labels $z_{i,c}$ follows standard MIL:
\begin{itemize}
    \item If $y_c=1$, then $\max_{i=1,...,N_b} \{z_{i,c}\} = 1$.
    \item If $y_c=0$, then $z_{i,c} = 0, \forall i \in \{1, ..., N_b\}$.
\end{itemize}

\subsubsection{Expectation-Maximization (EM) Framework}
We treat the patch labels $z_{i,c}$ as unobserved latent variables and optimize the ARM parameters $\theta$ using EM.

\paragraph{\textbf{E-step: Computing Smoothed Responsibilities $\hat{z}'_{i,c}$}}
The E-step aims to compute the expected value of the latent variables $z_{i,c}$ given the observed data and current parameters, i.e., the posterior probability $P(z_{i,c}=1 | y, b; \theta)$.

\textbf{Case 1: Negative Bag ($y_c=0$)}
From the MIL assumption, $z_{i,c}=0$ for all $i$. Thus, the posterior probability $P(z_{i,c}=1 | y_c=0, b; \theta) = 0$. Consequently, both the model-predicted responsibility $\hat{z}_{i,c}$ and the prior responsibility $\hat{z}^*_{i,c}$ are 0 for negative classes. The final smoothed responsibility is also zero:
\begin{equation}
\hat{z}'_{i,c} = 0 \quad \text{if } y_c = 0. \tag{C.1}
\end{equation}

\textbf{Case 2: Positive Bag ($y_c=1$)}
We need $P(z_{i,c}=1 | y_c=1, b; \theta)$. Applying Bayes' theorem leads to an intractable expression:
\begin{equation}
P(z_{i,c}=1 | y_c=1, b; \theta) = \frac{P(z_{i,c}=1 | b; \theta)}{1 - \prod_{j=1}^{N_b} (1 - P(z_{j,c}=1 | b; \theta))}, \tag{C.2}    
\end{equation}
where $P(z_{i,c}=1 | b; \theta)$ is often approximated by $\sigma(\tilde{S}_{i,c})$.

\textbf{Softmax Approximation:} We approximate the posterior distribution over patches within a positive bag using the Softmax function applied to the relevant scores.
The model-predicted responsibility $\hat{z}_{i,c}$ is computed using the current model's scores $\tilde{S}_{i,c}$:
\begin{equation}
\hat{z}_{i,c} = \begin{cases} \frac{\exp(\tilde{S}_{i,c} / \tau)}{\sum_{j=1}^{N_b} \exp(\tilde{S}_{j,c} / \tau)} & \text{if } y_c = 1 \\ 0 & \text{if } y_c = 0 \end{cases} \tag{C.3}
\end{equation}
Here, $\tau$ is a temperature hyperparameter.

The prior responsibility $\hat{z}^*_{i,c}$ is computed similarly using the original VLP scores $S^*_{i,c}$ and a temperature $\tau_{\text{prior}}$:
\begin{equation}
\hat{z}^*_{i,c} = \begin{cases} \frac{\exp(S^*_{i,c} / \tau_{\text{prior}})}{\sum_{j=1}^{N_b} \exp(S^*_{j,c} / \tau_{\text{prior}})} & \text{if } y_c = 1 \\ 0 & \text{if } y_c = 0 \end{cases} \tag{C.4}
\end{equation}

\textbf{Smoothing:} The final smoothed responsibility $\hat{z}'_{i,c}$ used in the M-step is a convex combination of the model-predicted and prior responsibilities:
\begin{equation}
    \hat{z}'_{i,c} = \lambda \hat{z}^*_{i,c} + (1-\lambda) \hat{z}_{i,c}, \tag{C.5}
\end{equation}
where $\lambda \in [0, 1]$ is the smoothing hyperparameter, typically annealed during training. Note that if $y_c=0$, both $\hat{z}_{i,c}=0$ and $\hat{z}^*_{i,c}=0$, correctly yielding $\hat{z}'_{i,c}=0$.

\paragraph{\textbf{M-step: Updating Model Parameters $\theta$}}
The M-step minimizes the negative expected complete-data log-likelihood using the smoothed responsibilities $\hat{z}'_{i,c}$ computed in the E-step. The objective function is equivalent to minimizing the sum of Binary Cross-Entropy (BCE) losses weighted by $\hat{z}'_{i,c}$:
\begin{equation}
\mathcal{L}_{\text{M-step}}(\theta; \hat{Z}') = -\sum_{b \in \mathcal{B}} \sum_{i=1}^{N_b} \sum_{c=1}^{C} \text{BCE}(\sigma(\tilde{S}_{i,c}), \hat{z}'_{i,c}), \tag{C.6}
\end{equation}
where $\text{BCE}(p, q) = -[q \log p + (1-q) \log(1-p)]$. Explicitly:
\begin{equation}
\begin{split}
\mathcal{L}_{\text{M-step}}(\theta; \hat{Z}') &= \sum_{b \in \mathcal{B}} \sum_{i=1}^{N_b} \sum_{c=1}^{C} \Big[ -\hat{z}'_{i,c} \log \sigma(\tilde{S}_{i,c}) \\
&\quad - (1-\hat{z}'_{i,c}) \log(1 - \sigma(\tilde{S}_{i,c})) \Big].
\end{split}
\tag{C.7}
\end{equation}

We now formulate the final objective by analyzing the loss contributions from positive ($c \in \mathcal{P}_b$) and negative ($c \in \mathcal{N}_b$) classes separately.

\textbf{Positive Classes ($c \in \mathcal{P}_b$):} The contribution from the BCE term for a single patch $i$ and positive class $c$ is $-[\hat{z}'_{i,c} \log \sigma(\tilde{S}_{i,c}) + (1-\hat{z}'_{i,c}) \log(1 - \sigma(\tilde{S}_{i,c}))]$. Following the simplification described in Sec.~3.3, we retain only the first part, focusing on maximizing the evidence for responsible patches:
\begin{equation}
    \mathcal{L}_{\text{pos\_term}}(i, c; \theta) = -\hat{z}'_{i,c} \log \sigma(\tilde{S}_{i,c}) \quad \text{for } c \in \mathcal{P}_b. \tag{C.8}
\end{equation}

\textbf{Negative Classes ($c \in \mathcal{N}_b$):} For negative classes, we established that the responsibility $\hat{z}'_{i,c} = 0$. Substituting $\hat{z}'_{i,c}=0$ into the full BCE term (Eq. C.7) gives:
\begin{equation}    
-[0 \cdot \log \sigma(\tilde{S}_{i,c}) + (1-0) \log(1 - \sigma(\tilde{S}_{i,c}))] = -\log(1 - \sigma(\tilde{S}_{i,c})),\notag
\end{equation}
Using the identity $1 - \sigma(s) = \sigma(-s)$, the contribution becomes:
\begin{equation}
    \mathcal{L}_{\text{neg\_term}}(i, c; \theta) = -\log \sigma(-\tilde{S}_{i,c}) \quad \text{for } c \in \mathcal{N}_b. \tag{C.9}
\end{equation}

\textbf{Hard Negative Mining (HNM):} For each image $b$ and negative class $c \in \mathcal{N}_b$, the model should ideally predict low scores for all patches. To focus training on the most erroneous predictions, we apply HNM. We identify the set of $K$ patches with the highest scores, $\mathcal{I}_{b,c}^{\text{hard}} = \text{TopkIdx}_j (\text{sg}[\tilde{S}_{j,c}], K)$, where $\text{sg}[\cdot]$ is the stop-gradient operator. The negative loss is then computed only over these hardest examples:
\begin{equation}
\mathcal{L}_{\text{neg-HNM}}(b, c; \theta) = \sum_{j \in \mathcal{I}_{b,c}^{\text{hard}}} [-\log \sigma(-\tilde{S}_{j,c})] . \tag{C.10}
\end{equation}

\subsubsection{Final WPS Loss Formulation}
Combining the positive loss from Eq. (C.8) and the hard negative loss from Eq. (C.10), we arrive at the final WPS loss objective to be minimized, aggregated over the batch $\mathcal{B}$:
\begin{align}
\mathcal{L}_\text{WPS}(\theta) =
\sum_{b \in \mathcal{B}} \Big(
&\underbrace{
-\sum_{c \in \mathcal{P}_b} \sum_{i=1}^{N_b}
\hat{z}'_{i,c}
\log \sigma(\tilde{S}_{i,c})
}_{\text{Weighted Positive Loss}} \notag\\
&\underbrace{
-\sum_{c \in \mathcal{N}_b}
\sum_{j \in \mathcal{I}_{b,c}^{\text{hard}}}
\log(\sigma(-\tilde{S}_{j,c}))
}_{\text{Hard Negative Loss}} \Big),
\tag{C.11}
\end{align}
where $\hat{z}'_{i,c} = \lambda \hat{z}^*_{i,c} + (1-\lambda) \hat{z}_{i,c}$ is the smoothed pseudo-label responsibility.

\subsection{Adaptive Transfer Module: GAT Implementation Details}
\label{app:gat_details}

This appendix provides further details on the Graph Attention Network (GAT) mechanism employed within the Adaptive Transfer Module (ATM), as introduced in Sec~3.4. Our implementation utilizes multi-head attention based on the principles of GATv2 \citep{gatv2} to facilitate dynamic and expressive adaptive information transfer across the Class Relationship Graph (CRG) $\mathcal{G}$.

\subsubsection{GATv2 Attention Mechanism}

Recall from Sec~3.4 that the ATM uses GAT layers to update class representations $h_l^{(c)}$ based on information aggregated from neighboring nodes $j \in \mathcal{N}(c) \cup \{c\}$ (including self-loops) in the CRG $\mathcal{G}$. The core of the adaptive transfer lies in computing the attention coefficients $\hat{\alpha}_{cj}$. We adopt the GATv2 formulation for computing the unnormalized attention score $e_{cj}$ between nodes $c$ and $j$ at layer $l$:
\begin{align}
        e_{cj} &= \mathbf{a}^\top \text{LeakyReLU}\left( \mathbf{W}_{\text{attn}} [h_l^{(c)} \Vert h_l^{(j)}] \right) \notag \\
    &= \mathbf{a}^\top \text{LeakyReLU}\left( \mathbf{W}_{\text{left}} h_l^{(c)} + \mathbf{W}_{\text{right}} h_l^{(j)} \right),
    \tag{C.12}
    \label{eq:app_gatv2_score}
\end{align}
where:
\begin{itemize}
    \item $h_l^{(c)}$ and $h_l^{(j)}$ are the feature representations of class nodes $c$ and $j$ at layer $l$.
    \item $\mathbf{W}_{\text{left}}$ and $\mathbf{W}_{\text{right}}$ are learnable weight matrices projecting the source and target node features, respectively, within the attention computation. $\mathbf{W}_{\text{attn}}$ represents the combined transformation.
    \item $[\cdot \Vert \cdot]$ denotes feature concatenation.
    \item $\mathbf{a}$ is a learnable weight vector that scores the transformed pair.
    \item LeakyReLU is the activation function, introducing non-linearity.
\end{itemize}
Critically, unlike the original GAT \citep{gat}, this GATv2 scoring function allows the attention weight $\hat{\alpha}_{cj}$ to depend on both $h_l^{(c)}$ and $h_l^{(j)}$ *after* their respective linear transformations ($\mathbf{W}_{\text{left}}, \mathbf{W}_{\text{right}}$), enabling a more dynamic and expressive attention mechanism where the ranking of neighbors can change based on the query node $c$.

The unnormalized scores $e_{cj}$ are then normalized across all neighbors of node $c$ (including itself) using the softmax function to obtain the final attention coefficients $\hat{\alpha}_{cj}$:
\begin{equation}
    \hat{\alpha}_{cj} = \text{softmax}_j(e_{cj}) = \frac{\exp(e_{cj})}{\sum_{k \in \mathcal{N}(c) \cup \{c\}} \exp(e_{ck})}
    \label{eq:app_gat_softmax}. \tag{C.13}
\end{equation}
These coefficients $\hat{\alpha}_{cj}$ represent the adaptively learned importance of node $j$'s features for updating node $c$'s representation at layer $l$.

\subsection{Multi-Head Attention}
To enhance the model's capacity and stabilize the learning process, we employ multi-head attention. Let $M$ be the number of independent attention heads.
Each head $m \in \{1, \dots, M\}$ performs the GATv2 attention computation described above using its own set of learnable parameters: $\mathbf{W}_{\text{left}}^{(m)}$, $\mathbf{W}_{\text{right}}^{(m)}$, and $\mathbf{a}^{(m)}$. This results in head-specific attention coefficients $\hat{\alpha}_{cj}^{(m)}$.

Each head $m$ then computes an aggregated feature vector by performing a weighted sum of the linearly transformed neighbor features $h_l^{(j)}$, using its learned attention coefficients $\hat{\alpha}_{cj}^{(m)}$. The linear transformation within the aggregation is performed by a head-specific weight matrix $\mathbf{W}_l^{(m)}$:
\begin{equation}
    h_{\text{aggr}}^{(c, m)} = \sigma_{\text{GAT}} \left( \sum_{j \in \mathcal{N}(c) \cup \{c\}} \hat{\alpha}_{cj}^{(m)} \mathbf{W}_l^{(m)} h_l^{(j)} \right),
    \label{eq:app_gat_head_aggr} \tag{C.14}
\end{equation}
where $\sigma_{\text{GAT}}$ is an activation function (e.g., ELU or LeakyReLU) applied after the aggregation within each head.

The outputs from all $M$ heads are typically combined. In our implementation, we concatenate the resulting feature vectors:
\begin{equation}
    h_{\text{concat}}^{(c)} = \mathop{\Vert}\limits_{m=1}^M h_{\text{aggr}}^{(c, m)}.
    \label{eq:app_gat_concat} \tag{C.15}
\end{equation}

\subsubsection{Final Node Update}
The concatenated feature vector $h_{\text{concat}}^{(c)}$ represents the aggregated information from all neighbors across all attention heads. This aggregated information is then used to update the node's representation. Incorporating the residual connection, as shown in Eq.~5 in the main text, the final update rule for node $c$ from layer $l$ to $l+1$ is:
\begin{equation}
    h_{l+1}^{(c)} = h_{\text{concat}}^{(c)} + h_l^{(c)}
    \label{eq:app_gat_update}. \tag{C.16}
\end{equation}
In some architectures, a final linear transformation and/or normalization layer might be applied to $h_{\text{concat}}^{(c)}$ before the residual addition, particularly if the concatenated dimension needs to be adjusted back to the original feature dimension.

This multi-head GATv2 mechanism is applied iteratively for $l_G$ layers within both the Text-ATM and the MM-ATM stages of our DART framework, enabling the adaptive leveraging of relational knowledge encoded in the CRG for enhancing both linguistic and multi-modal class representations.

\subsection{LLM-based Class Relationship Graph Construction}
\label{app:crg_construction}

The Adaptive Transfer Module (ATM) leverages a pre-constructed Class Relationship Graph (CRG), denoted $\mathcal{G}=(V, E)$, to facilitate adaptive inter-class information transfer using Graph Attention Networks (GATs). Constructing an effective CRG is crucial. While the GAT mechanism adaptively weights neighbors, defining the initial graph structure—which nodes are considered neighbors—significantly impacts performance and efficiency.

Connecting all potential classes (nodes) in the graph, especially with a large vocabulary encompassing both seen and unseen classes, presents challenges. Dense connectivity can lead to computational inefficiency and potential over-smoothing of node features during message passing \citep{gnnsmooth}, where representations become indistinguishable. Furthermore, indiscriminately connecting classes may introduce noise from irrelevant nodes, hindering effective transfer. More fundamentally, relying on relationships derived solely from training set co-occurrence statistics is untenable in the Open-Vocabulary Multi-Label Recognition (OV-MLR) setting, as such data is unavailable for unseen classes. Simple semantic similarity from VLP embeddings might also be insufficient, capturing only one facet of complex inter-class dependencies.

Our goal is therefore to construct a CRG that is both informative and computationally manageable, encoding diverse, generalizable relationships beyond dataset statistics or basic similarity. Intuitively, not every category provides useful context for understanding another; for instance, knowing about a "computer" is unlikely to aid the recognition of a "giraffe". To capture meaningful connections reflecting real-world knowledge, we leverage the reasoning capabilities and extensive knowledge base of Large Language Models (LLMs). We identify several types of inter-class relationships pertinent to visual scene understanding:

\begin{itemize}
    \item \textbf{Synonymy/Similarity}: Conceptual similarity (e.g., "dog" and "puppy").
    \item \textbf{Is-a/Hypernymy/Hyponymy}: Hierarchical relationships (e.g., "orchid" is a type of "flower").
    \item \textbf{Functional Relationship}: Association based on use or interaction (e.g., "pandas" are often found in a "zoo").
    \item \textbf{Co-occurrence}: Frequent appearance in similar contexts (e.g., "fish" and "reef").
    \item \textbf{Part-Whole Relationship}: Compositional relationships (e.g., "brown" as a property/part of "bear").
\end{itemize}

Based on these relationship heuristics, we design carefully engineered prompts (illustrated in the \textbf{black box}) to query an LLM. As detailed in Sec.~3.4, these prompts guide the LLM to identify related classes for a given target class, potentially requesting justifications and relevance scores to enhance reliability. By aggregating the high-confidence relationships mined across the entire vocabulary (including both seen and unseen classes), we form the edges $E$ of the CRG. This process yields a graph $\mathcal{G}$ that explicitly models diverse inter-category dependencies based on broad world knowledge, providing a robust structure for the GATs in the ATM to perform adaptive information transfer, crucial for effective OV-MLR.

\begin{tcolorbox}[breakable, colback=white, colframe=black, title=Prompt]

Based on the following list of known categories, please identify all categories that have a direct relationship with the new category \textbf{\{\texttt{New Category}\}}. For each related category, provide the type of relationship, the association strength (\textbf{High}, \textbf{Medium}, \textbf{Low}), and an explanation.

\subsubsection*{Types of Relationships}

\begin{enumerate}[label=\arabic*.]
    \item \textbf{Synonymy/Similarity}: Two categories are conceptually very similar or synonymous.
    \item \textbf{Is-a/Hypernym}: One category is a superordinate or subordinate concept of the other.
    \item \textbf{Functional Relationship}: The function or use of one category is related to the other.
    \item \textbf{Co-occurrence}: Two categories often appear in the same context or environment.
    \item \textbf{Part-Whole Relationship}: One category is a component of the other.
\end{enumerate}

\subsubsection*{Instructions}

Please provide the information for each relevant category in the following format:

\textbf{Related Category [Number]: [Category Name]}
\begin{itemize}[leftmargin=*]
    \item \textbf{Type of Relationship}: [Relationship Type]
    \item \textbf{Association Strength}: High / Medium / Low
    \item \textbf{Explanation}: [Brief explanation of the relationship and the reason for the assigned strength]
\end{itemize}

\subsubsection*{Example}

\textbf{New Category}: Nature

\textbf{List of Seen Categories}:
natural, fauna, wildlife, flora, scenic, outdoors, cliff, blossoms, insect, wild, plant, scenery, blooms, gardens, landscapes

\textbf{Example Output}:

\begin{itemize}[leftmargin=*, label={}]
    \item \textbf{Related Category 1: natural}
        \begin{itemize}[leftmargin=*]
            \item \textbf{Type of Relationship}: Synonymy/Similarity
            \item \textbf{Association Strength}: High
            \item \textbf{Explanation}: ``Natural'' is conceptually very similar to ``nature'' as both refer to elements of the physical world not created by humans.
        \end{itemize}

    \item \textbf{Related Category 2: fauna}
        \begin{itemize}[leftmargin=*]
            \item \textbf{Type of Relationship}: Is-a/Hypernym
            \item \textbf{Association Strength}: High
            \item \textbf{Explanation}: ``Fauna'' represents the animal life of a region, which is a fundamental part of ``nature''.
        \end{itemize}

    \item \textbf{...}
\end{itemize}

Using the format and example provided above, identify all categories from the list of known categories that have a direct relationship with the new category \textbf{\{\texttt{New Category}\}}. For each related category, specify:

\textbf{List of Seen Categories}:

\{\texttt{List of Seen Categories}\}

Focus on associations that would be most relevant for understanding or classifying \textbf{\{\texttt{New Category}\}} within this domain.
\end{tcolorbox}

\vspace{5ex}

\section{More Experiments}

\subsection{Model Complexity and Efficiency Analysis}
\label{sec:appendix_complexity}

We provide a comprehensive analysis of the model complexity, comparing DART with CLIP and a strong baseline, MKT. All experiments were benchmarked on a single NVIDIA RTX 3090 GPU with an input image resolution of 224x224. As shown in Tab.~\ref{tab:complexity}, MKT's reliance on an additional CLIP model for knowledge distillation results in nearly double the total parameters compared to our method. In contrast, DART features a more lightweight design with fewer trainable parameters and significantly faster training speeds. While its inference time is slightly higher than MKT's, it remains highly efficient, demonstrating a favorable trade-off between performance and computational cost.

\vspace{1ex}
\begin{table}[h]
\centering
\begin{tabular}{lcccc}
\toprule
\textbf{Model} & \textbf{Param} & \textbf{Trainable} & \textbf{Inf. Time} & \textbf{Train Time} \\
\midrule
CLIP           & 86.7           & /                  & 5.5                & /                   \\
MKT            & 175.1          & 7.5                & 7.2                & 172.5               \\
\textbf{Ours}  & \textbf{92.7}  & \textbf{6.5}       & 9.6                & \textbf{106.8}      \\
\bottomrule
\end{tabular}
\caption{Comparison of model complexity and efficiency. 
Abbreviations: Param (total parameters in millions), Trainable (trainable parameters in millions), Inf. Time (inference time in milliseconds), Train Time (training time per batch in milliseconds). Our method demonstrates a lightweight design with fewer trainable parameters and faster training time compared to MKT.}
\label{tab:complexity}
\end{table}

\begin{figure*}[t!]
    \centering
    \begin{subfigure}[b]{0.33\textwidth}
        \includegraphics[width=\linewidth]{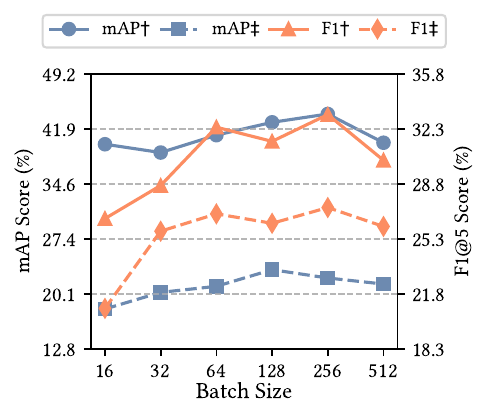}
        \subcaption{Effect of batch size}
        \label{fig:a_batch}
    \end{subfigure}
    \begin{subfigure}[b]{0.33\textwidth}
        \includegraphics[width=\linewidth]{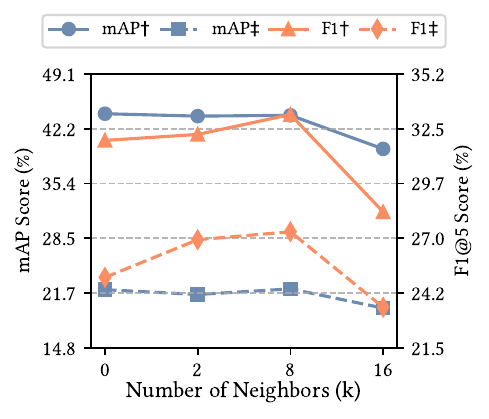}
        \subcaption{Effect of neighbor num on CRG.}
        \label{fig:b_cgrg}
    \end{subfigure}
    \begin{subfigure}[b]{0.33\textwidth}
        \includegraphics[width=\linewidth]{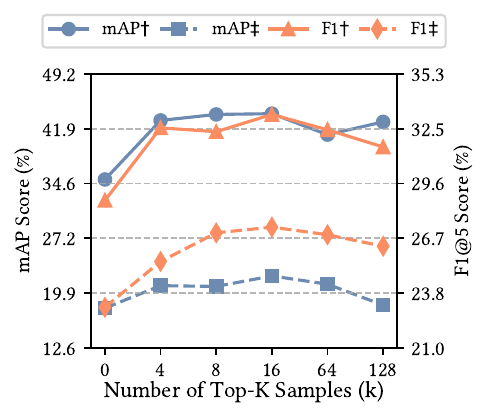}
        \subcaption{Effect of top-k on ARM.}
        \label{fig:c_topk}
    \end{subfigure}
    \caption{Ablation experiments on our components. ($\dagger$: ZSL result; $\ddagger$: GZSL result)}
    \label{fig:loss}
\end{figure*}

\subsection{More Ablation Study}

\noindent\textbf{Impact of the Hyperparameter $\lambda$ in WPS}.
The hyperparameter $\lambda$ in our Weakly Supervised Patch (WPS) loss is crucial for balancing the learning objective between the guidance from the frozen VLP's scores and our model's refined predictions. We conducted an ablation study to analyze its impact, with results presented in Tab.~\ref{tab:lambda}. Relying solely on the model's refined scores ($\lambda=0$) or the original VLP scores ($\lambda=1$) leads to suboptimal performance. Our scheduled approach, which creates a curriculum that gradually shifts focus from the VLP's guidance to the model's own predictions, achieves the best results. This demonstrates that our strategy effectively stabilizes training by balancing knowledge preservation and adaptive learning.

\noindent\textbf{Analysis of Patch Selection Strategy}.
We investigated whether a dynamic patch selection strategy could outperform our Top-K approach. We experimented with a dynamic threshold-based selection (patch score > 0.5) during training and inference. The results in Tab.~\ref{tab:topk_strategy} show that using a dynamic threshold during training is unstable, particularly in the early epochs, and results in a significant performance drop (-4.8\% mAP@ZSL). Using a dynamic threshold for inference after training with our standard Top-K method yields results comparable to our original approach. Therefore, our experiments confirm that the fixed Top-K strategy is a robust and well-justified design choice that provides crucial training stability for superior final model performance.

\noindent\textbf{Ablation on ATM Components}. Tab.~\ref{tab:atm} ablates the two stages of the ATM. Applying only Text-ATM or only MM-ATM improves performance over the baseline without ATM ('None'), indicating the benefit of relational transfer in both linguistic and multi-modal domains. However, the full ATM employing both stages sequentially achieves the best overall results, particularly boosting F1 scores (e.g., ZSL F1@3 of 36.1 vs. 35.9/34.4; GZSL F1@5 of 27.3 vs. 25.7/26.0). This confirms the synergistic value of first enhancing text embeddings and then performing relational interaction on the fused multi-modal features.

\vspace{1ex}
\begin{table}[h]
\centering
\begin{tabular}{lcc}
\toprule
\textbf{$\lambda$} & \textbf{mAP@ZSL} & \textbf{mAP@GZSL} \\
\midrule
0                & 36.7             & 16.9              \\
0.5              & 42.3             & 21.8              \\
1                & 40.2             & 20.9              \\
\textbf{Scheduled} & \textbf{43.9}    & \textbf{22.2}     \\
\bottomrule
\end{tabular}
\caption{Ablation study on the $\lambda$ parameter in WPS loss on NUS-WIDE. The scheduled approach yields the best performance on both ZSL and GZSL settings.}
\label{tab:lambda}
\end{table}

\begin{table}[h]
\centering
\setlength{\tabcolsep}{2pt}
\begin{tabular}{lcc}
\toprule
\textbf{Selection Strategy} & \textbf{mAP@ZSL} & \textbf{mAP@GZSL} \\
\midrule
Dynamic Threshold (Train)   & 39.1             & 19.5              \\
Top-K (Train) + Threshold (Infer) & 42.1             & 22.3              \\
\textbf{Top-K (Ours)}       & \textbf{43.9}    & \textbf{22.2}     \\
\bottomrule
\end{tabular}
\caption{Ablation study on the patch selection strategy on NUS-WIDE. Our Top-K strategy provides essential training stability and achieves the best performance.}
\label{tab:topk_strategy}
\end{table}

\noindent\textbf{Ablation on Batch Size.} As shown in Fig.~\ref{fig:a_batch}, we conducted experiments to evaluate the impact of batch size on model performance. We observed a general trend where increasing the batch size from smaller values led to improved results across key metrics. Performance peaked when using a batch size of 256. However, further increasing the batch size to 512 resulted in a slight degradation in performance. This suggests that while larger batch sizes can offer more stable gradient estimates and potentially better convergence up to a point, excessively large batches might hinder generalization or require adjustments to optimization parameters (like learning rate) to maintain optimal performance within our training framework. Therefore, a batch size of 256 was chosen as the optimal setting for our experiments.

\noindent\textbf{Ablation on CRG Neighbor Count}. As shown in Fig.~\ref{fig:b_cgrg}, we investigated the effect of varying the number of neighbors ($N$) selected for each class when constructing the CRG based on LLM-mined relationships. Our experiments showed that performance generally increased as $N$ grew from small values, indicating the benefit of incorporating more relational context. However, this trend reversed after a certain point; using too many neighbors introduced noise and potentially diluted the influence of the most relevant connections. We found that selecting the top $N=8$ neighbors for each class yielded the best performance in our experiments.

\noindent\textbf{Ablation on Top-K Patch Selection}. As shown in Fig.~\ref{fig:c_topk}, we also explored the impact of the number of top-scoring patches ($K$) aggregated to form the class-attentive visual features $x_\text{vis}^{(c)}$, based on the scores $\tilde{S}$ derived from the ARM's refined features. Selecting too few patches ($K$ too small) might miss important visual evidence, while selecting too many ($K$ too large) could incorporate irrelevant background or noisy patches. Our results indicated that a moderate value of $K=16$ achieved the optimal balance, leading to the best performance by effectively capturing salient object features while mitigating noise.

\begin{table}[t!]
    \setlength{\belowrulesep}{1pt}
    \centering
    \begin{tabular}{l ccc ccc}
        \toprule[.75pt]
        \multirow{2}{*}{\bf{Relation}} & \multicolumn{3}{c}{\bf{ZSL}} & \multicolumn{3}{c}{\bf{GSL}} \\
         & F1@3 & F1@5 & mAP & F1@3 & F1@5 & mAP\\
         \midrule
         None       & 34.0 & 31.9 & 44.2 & 22.7 & 25.0 & 22.1 \\
         Text-ATM   & 35.9 & 31.8 & 43.6 & 23.8 & 25.7 & 21.9 \\
         MM-ATM     & 34.4 & 32.4 & 44.0 & 22.3 & 26.0 & 22.0 \\
         ATM        & 36.1 & 33.2 & 43.9 & 23.8 & 27.3 & 22.2 \\
         \bottomrule[.75pt]
    \end{tabular}
    \caption{Ablation result (\%) of ATM.}
    \label{tab:atm}
\end{table}

\begin{figure}[!t]
    \centering
    \includegraphics[width=\linewidth]{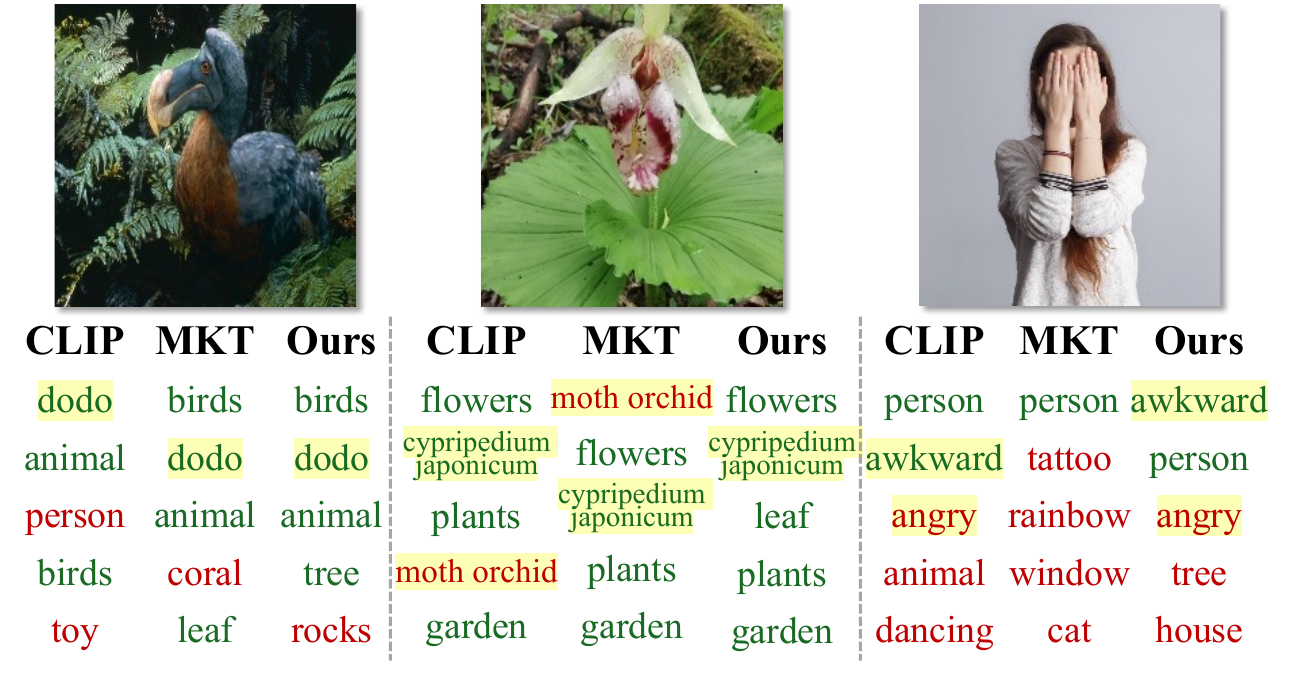}
    \caption{The top-5 category prediction results of each model in the open-vocabulary setting.}
    \label{fig:ov}
\end{figure}

\begin{figure*}[!t]
    \centering
    \subcaptionbox{
        The co-occurrence relation graph\label{fig:g-cooc}
    }{
        \includegraphics[width=.46\linewidth]{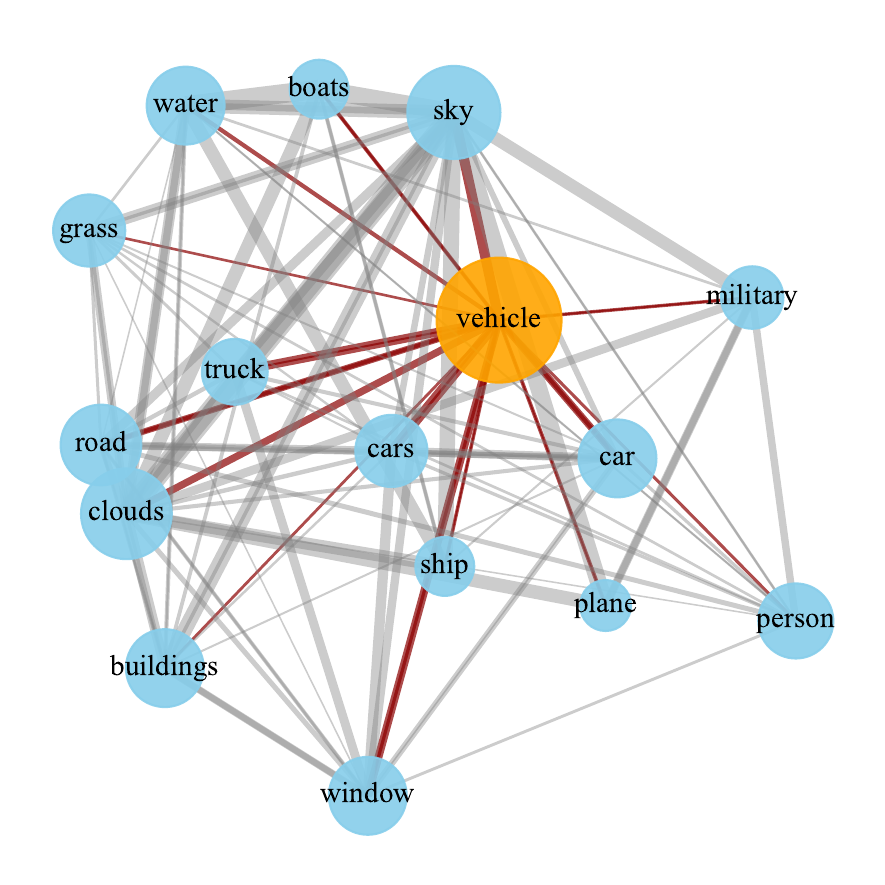}\vspace{-5pt}
    }\subcaptionbox{
        The similarity relation graph\label{fig:g-sim}
    }{
        \includegraphics[width=.46\linewidth]{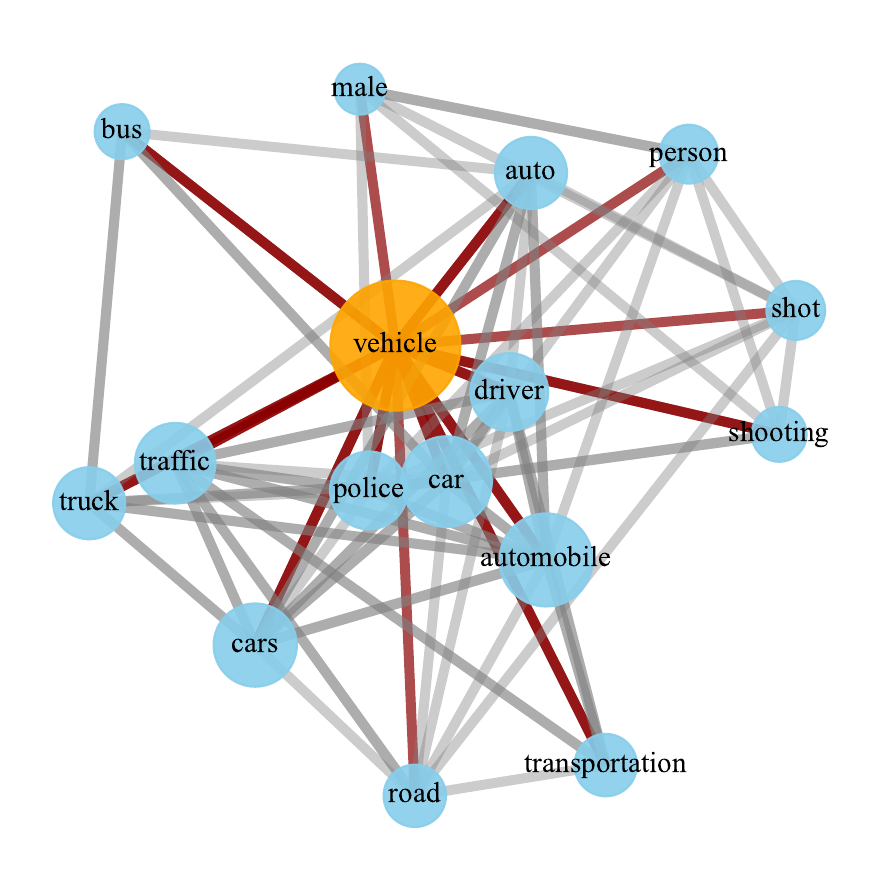}\vspace{-5pt}
    }
    \subcaptionbox{
        The LLM mining relation graph\label{fig:g-llm}
    }{
        \includegraphics[width=.46\linewidth]{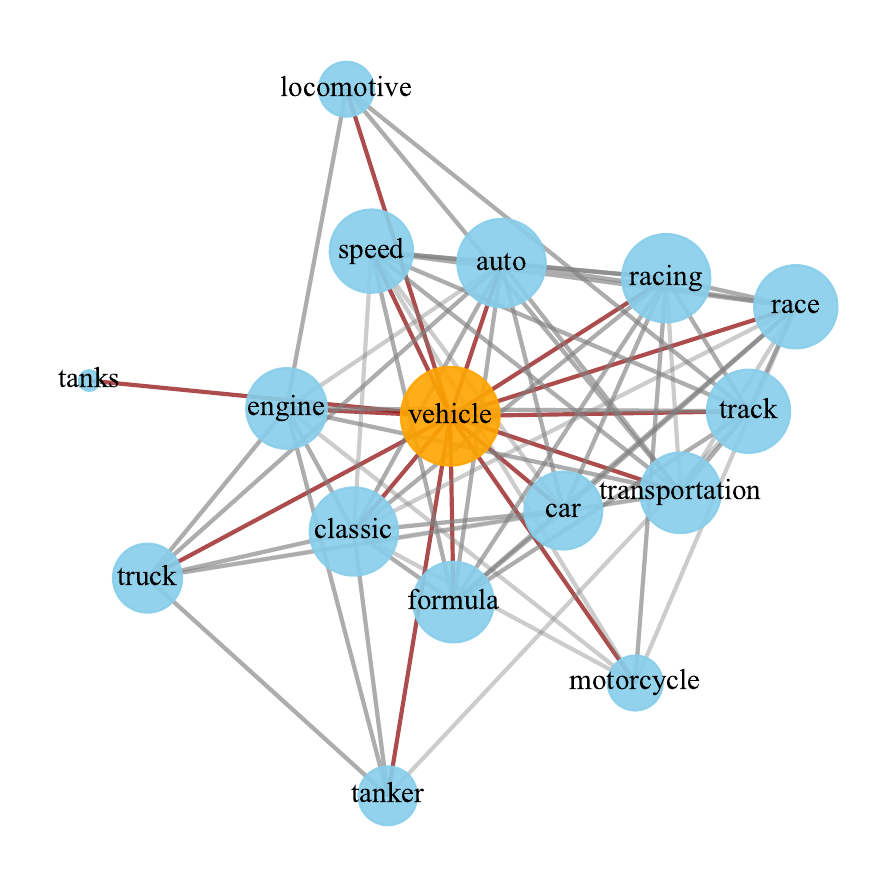}\vspace{-5pt}
    }
    \caption{
        Visualization of Diverse Inter-Class Relationship Graphs: (a) Co-occurrence Probability, (b) Similarity, and (c) LLM Mining. Orange nodes represent the target category, while blue nodes denote other categories. Red edges indicate information transmission from other categories to the target category, and gray edges represent information transmission among other categories. Node sizes correspond to the number of adjacent nodes, and edge widths reflect edge weights.
    }
    \label{fig:graph-relation}
    \vspace{7pt}
\end{figure*}

\begin{figure*}[!t]
    \centering
    \includegraphics[width=0.98\linewidth]{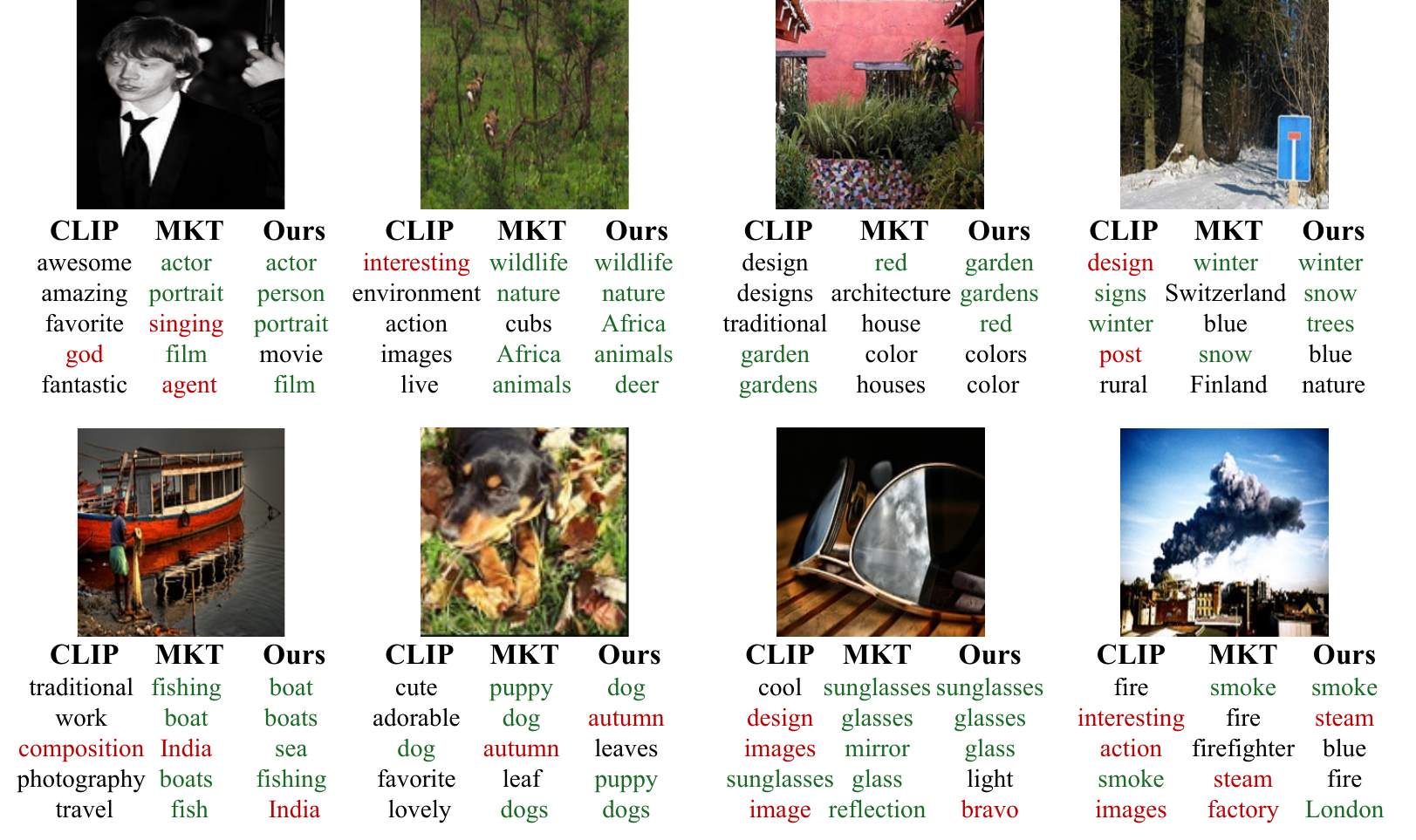}
    \caption{Comparison of Predictions among CLIP\cite{clip}, MKT\cite{mkt}, and Our Model. Green denotes correct predictions present in the dataset's ground truth labels, red denotes incorrect predictions, and black denotes predictions not present in the ground truth but are actually correct from a human perspective.}
    \label{fig:example}
    \vspace{5pt}
\end{figure*}

\noindent\textbf{Evaluation of Open-Vocabulary Recognition.}
To evaluate the open-vocabulary capabilities, we select novel images and categories that are absent from the evaluation dataset and rarer as well as more challenging. The results shown in Fig.~\ref{fig:ov} indicate that MKT's classification performance is inferior to that of CLIP, potentially due to bias introduced during training. In contrast, our method shows superior recognition ability in this open-vocabulary setting, effectively utilizing information from seen categories to enhance the recognition of novel ones, thus demonstrating significant potential.

\subsection{Visualization of Various Category Relation}
We visualize the related categories as shown in Fig.~\ref{fig:graph-relation}. Due to the large number of categories in the dataset, visualizing all category relationships becomes indistinguishable. Therefore, we focus on visualizing the inter-category relationship graph for a target category, ``vehicle". For the co-occurrence probability graph, we directly count the occurrence frequency of all categories in the dataset and calculate the co-occurrence probabilities. We select the categories with the highest co-occurrence probabilities as adjacent categories, and the edge weights correspond to the co-occurrence probabilities. Similarly, in the similarity relationship graph, we select the categories with the highest textual feature similarity as adjacent categories, with edge weights representing the similarity scores. In the inter-category relationship graph mined by the LLM, we query the LLM to obtain related categories to construct the inter-class relationship graph without edge weights, as it is challenging to accurately obtain quantitative relationship values between categories through the LLM.

As shown in Fig.~\ref{fig:graph-relation}\subref{fig:g-cooc}, the co-occurrence probability graph, a significant portion of the categories contribute to the understanding of ``vehicle", such as ``truck" and ``cars". However, there are also common and mundane categories like ``sky", which have high co-occurrence probabilities with the target category due to their frequent appearance in images. In reality, these categories do not positively contribute to the recognition of ``vehicle" and may even have a negative impact.

In the similarity relationship graph depicted in Fig.~\ref{fig:graph-relation}\subref{fig:g-sim}, the textual feature similarities exhibit low variance, resulting in almost identical edge weights and low discriminative power. This leads to the inclusion of additional categories beyond those with the highest similarity, such as ``male" and ``person", which have high similarity rankings but little relevance to the target category. Moreover, the similarity relationships lack some semantically dissimilar categories that are beneficial for recognition, such as ``race".

Fig.~\ref{fig:graph-relation}\subref{fig:g-llm} presents the inter-category relationship graph mined by the LLM. It is evident that the related categories contribute positively to the recognition of the target category while avoiding the inclusion of mundane categories introduced by co-occurrence probabilities and irrelevant categories introduced by similarity scores.

\subsection{More Qualitative Analysis}

To demonstrate the superiority of our model, Fig.~\ref{fig:example} presents additional prediction results compared with other methods on the NUS-WIDE dataset. As shown in the figure, due to label limitations in the dataset, black labels indicate categories that do not appear in the ground truth but can be inferred to be present in the image. This implies that even when our model's recognition results are not true positives, it still exhibits significant understanding and recognition capabilities. By adaptively leveraging both intra-class and inter-class semantic information, our model effectively enhances the accuracy and stability of open-vocabulary multi-label recognition.

\end{document}